\documentclass[journal]{IEEEtran}
\usepackage{times}
\usepackage{epsfig}
\usepackage{graphicx}
\usepackage{subcaption}
\usepackage{amsmath}
\usepackage{amssymb}
\usepackage{booktabs}
\usepackage{makecell}
\usepackage{dblfloatfix}
\usepackage{supertabular}
\usepackage{array}
\usepackage{url}
\usepackage{color}
\newcommand\blfootnote[1]{%
  \begingroup
  \renewcommand\thefootnote{}\footnote{#1}%
  \addtocounter{footnote}{-1}%
  \endgroup
}

\title{What and Where:  A Context-based Recommendation System for Object Insertion}

\author{Song-Hai~Zhang*~\IEEEmembership{Member,~IEEE,}
        Zhengping~Zhou*,
        Bin~Liu, Xin~Dong, Dun~Liang,
        Peter~Hall~\IEEEmembership{Member,~IEEE,}
        Shi-Min~Hu,~\IEEEmembership{Senior~Member,~IEEE,}
}

\IEEEoverridecommandlockouts
\begin{document}
        
\twocolumn[{%
\renewcommand\twocolumn[1][]{#1}%
\maketitle
\begin{center}
    \centering
    \includegraphics[width=.85\textwidth]{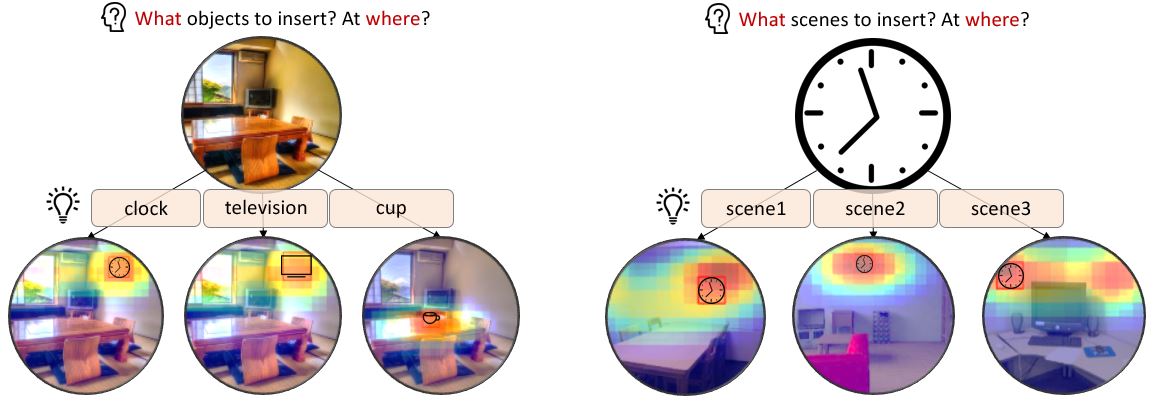}
    \captionof{figure}{This paper proposes a novel research topic consisting of two dual tasks: 1) Given a scene, recommend objects to insert (left); 2) Given an object category, retrieve suitable background scenes (right).}
\end{center}%
}]

\IEEEpeerreviewmaketitle

\begin{abstract}
In this work, we propose a novel topic consisting of two dual tasks: 1) given a scene, recommend objects to insert; 2) given an object category, retrieve suitable background scenes. A bounding box for the inserted object is predicted in both tasks, which helps downstream applications such as semi-automated advertising and video composition. The major challenge lies in the fact that the target object is neither present nor localized at test time, whereas available datasets only provide scenes with existing objects. To tackle this problem, we build an unsupervised algorithm based on object-level contexts, which explicitly models the joint probability distribution of object categories and bounding boxes with a Gaussian mixture model. Experiments on our newly annotated test set demonstrate that our system outperforms existing baselines on all subtasks, and do so under a unified framework. Our contribution promises future extensions and applications.

\blfootnote{

*\textit{Equal contribution.}

Song-Hai~Zhang, Zhengping~Zhou, Bin~Liu, Xin~Dong, Dun~Liang and Shi-Min~Hu are with the Department of Computer Science and Technology, Tsinghua University,
Beijing 100084, China.

Peter~Hall is with the Department of Computer Science Media Technology Research Center,
University of Bath, Bath BA2 7AY, United Kingdom.

Manuscript received Mon. Date, Year; revised Mon. Date, Year.

}

\end{abstract}

\IEEEpeerreviewmaketitle

\section{Introduction} \label{sec:intro}

Our goal is to build a bidirectional recommendation system  \cite{Ricci2010,wiki} that performs two tasks under a unified framework:
\begin{enumerate}
\item \textbf{Object Recommendation}: For a given scene, recommend a sorted list of \textit{categories} and \textit{bounding boxes} for insertable objects;
\item \textbf{Scene Retrieval}: For a given object category, retrieve a sorted list of suitable \textit{background} scenes and corresponding \textit{bounding boxes} for insertion.
\end{enumerate}

The motivation for the two  tasks stems from the bilateral collaboration between \textit{media owners} and \textit{advertisers} in the advertising industry. Some media owners make profits by offering paid promotion  \cite{youtube}, while many advertisers pay media owners for product placement \cite{mirriad}. This collaboration pattern reflects the mutual requirement, from which we distill the novel research topic of \textit{dual recommendation for object insertion}.  

Consider a typical collaborative workflow between a media owner and an advertising artist consisting of three phases:
\begin{enumerate}
\item Matching: The media owner determines \textit{what} kind of products are insertable, while the advertiser determines \textit{what} kind of background scenes are suitable. Both of them, in this process, also consider \textit{where}  an insertion might potentially happen;
\item Negotiation: They contact each other and confirm \textit{what and where} after negotiation;
\item Insertion: Post-process the media to perform the actual insertion.
\end{enumerate}

In this work, both of the above tasks aim underpin phases 1 and 2, but neither include a fully automatic solution for phase 3.  Analogously, the key idea here is to automatically make \textit{recommendations} rather than make \textit{decisions} for the user. We do not perform automatic segment selection or insertion, because in practice the inserted object will be brand-specific and the final decision depends upon the personal opinions of the advertiser \cite{lauda2010}. Nonetheless, for illustration purpose only, we use \textit{manually} selected, yet \textit{automatically} pasted object segments for cases presented in this paper, which demonstrates our system's ability to make reasonable recommendations on categories and bounding boxes.

The advantage of our system is three-fold. First, we provide constructive ideas for designers: the \textbf{object recommendation} task can be especially useful for sponsored media platforms, which may profit by making recommendations to media owners. Second, the \textbf{scene retrieval} task provides a specialized search engine that  is capable of retrieving images, given an object, that goes beyond previous content-based image retrieval systems \cite{DBLP:conf/cvpr/JohnsonKSLSBL15,7360966,7935507}.  Future applications include advertiser-oriented search engines, or matching services for designer websites. Third, the \textbf{bounding boxes} predicted for both tasks further makes the recommendation concrete and visualizable. As we will show in our experiments, this not only enables  applications such as automatic preview  over a gallery of target segments, but also may assist designers with a  heatmap as hint to  users.

Specifically, our contributions are:
\begin{enumerate}
\item We are the first, to the best of our knowledge, to propose \textit{dual recommendation for object insertion} as a new research topic;
\item We develop an unsupervised algorithm (Sect. \ref{sec:method}) based on object-level context \cite{rabinovich2007}, which explicitly models the joint probability distribution of object category and bounding box;
\item We establish a newly annotated test set (Sect. \ref{sec:dataset}), and introduce task-specific metrics for automatic quantitative evaluation (Sect. \ref{sec:exp});
\item We outperform existing baselines on all subtasks under a unified framework, as demonstrated by both quantitative and qualitative results (Sect. \ref{sec:exp}). 
\end{enumerate}

\section{Related Work}

Although there are no related works that directly addresses exactly the same topic, we can still borrow ideas from previous arts on related tasks.

\textbf{Object Recognition. } The family of recognition tasks include image classification \cite{resnet,imagenet_classification,going_deeper}, object detection \cite{DBLP:journals/corr/RenHG015,rich_feature, ssd}, weakly supervised object detection \cite{learning_deep,wsddn,contextlocnet} and semantic segmentation \cite{DBLP:journals/corr/HeGDG17,fully_conv,parsenet}.

Generally, the \textit{appearance} of the target object is given, and the expected output is either the \textit{category} (image classification), or the \textit{location} (weakly supervised object detection), or both (object detection, semantic segmentation).  Our \textbf{object recommendation} task shares the similar output of both \textit{category} and \textit{location}, but there are two key differences: (i) the \textit{appearance} of the target object is unknown in our task, for the object is even not present at the scene; (ii) the expected outputs for both \textit{category} and \textit{location} are not unique, for there may be multiple objects suitable for the same scene with multiple reasonable placements. 

In this work, we build our system upon the recently proposed state-of-the-art object detector, Faster R-CNN \cite{DBLP:journals/corr/RenHG015}. The basic idea is to seek evidence from other existing objects in the scene, which requires object detection as a basic building block. We also extend the expected output from a single \textit{category} and a single \textit{location} to lists of each, to allow multiple acceptable results in an information retrieval (IR) fashion.

\textbf{Image Retrieval. } Image retrieval tasks aim to retrieve a list of relevant images based on keywords \cite{DBLP:conf/cvpr/JohnsonKSLSBL15}, example images \cite{7935507}, or even other abstract concepts such as sketches or layouts \cite{recent_advance}. 

Generally, some \textit{attributes} (topic, features, color, layout, {\em etc.}) are known about the target image, and the expected output is a list of images that satisfies these conditions. Our \textbf{scene retrieval} task is distinct to this family of tasks because  our query object is \textit{not} generally present in the scene. Neither is it  an attribute possessed by the target image. 
Nonetheless, we share a similar idea as the retrieval systems in two aspects:  First, we adopt the similar expected output as a ranked list, and employ the metric, normalized discounted cumulative gain (nDCG) , as is widely used in previous retrieval tasks; Second, similar to content-based image retrieval systems  \cite{DBLP:conf/cvpr/JohnsonKSLSBL15,7360966,7935507}, we also utilize the known information of the image, typically the categories and locations of the existing objects. 

\textbf{Image Composition. } Our work aims to provide inspirations for object insertion, which has a close relationship to image composition. Some works focus on interactive editing, for instance, \cite{patchnet} builds an interactive library-based editing tool. It enables users to draw rough sketches, leading to plausible composite images incorporated with retrieved patches; Some other works focus on automatic completion, with image in-painting as one of the most notable research topics \cite{context_encoder, generative_inpainting}. These works aim to restore the removed region of an image, typically with neural networks that exploit the context. Our system is unique in two aspects: 1) We neither take the user's sketches as input, nor require a masked region as location hint; 2) We do not take ``plausible'' as our final goal, because our motivation is to do recommendations, rather than make decisions, as explained in Sect. \ref{sec:intro} .

Closest to our work is the automatic person composition task proposed by \cite{DBLP:journals/corr/TanBCOB17}, which establishes a fully automatic pipeline for incorporating a person into a scene. This pipeline consists of two stages: 1) location prediction; 2) segment retrieval. Though our system is different from this work, in that we do not perform segment retrieval; while it could not make recommendations on categories or scenes. We compare our system's performance on bounding box prediction with the first stage of this work, and report both quantitative and qualitative results. 

\section{Method} \label{sec:method}

In this section, first, we decompose the two tasks into three subtasks with probabilistic formulations, which we derive from the same joint probability distribution. Furthermore, we present an algorithm that models object-level context with a Gaussian mixture model (GMM), which leads to an approximation for the joint distribution. Finally, we report implementation details and per-image runtime.

\subsection{Problem Formulation} \label{sec:probform}

Given a set of candidate object categories $\cal C$, a set of scene images $\cal I$, and a set of candidate bounding boxes $\mathcal{B}_I$ for each specific image $I$, we further break the two tasks introduced in Sect. \ref{sec:intro} into the following three subtasks:
\begin{enumerate}
\item \textbf{Object Recommendation}: for a given image $I \in \mathcal{I}$, rank all candidate categories $C \in \mathcal{C}$ by $P(C|I)$; 
\item \textbf{Scene Retrieval}: for a given object category $C \in \mathcal{C}$, rank all candidate images $I \in \mathcal{I}$ by $P(I|C)$;
\item \textbf{Bounding Box Prediction}: for a given image $I \in \mathcal{I}$ and an object category $C \in \mathcal{C}$, rank all candidate bounding boxes $B \in \mathcal{B}_I$ by $P(B|C,I)$.
\end{enumerate}
We show that all of the three subtasks can be solved from the same joint probability distribution $P(B,C|I)$. The basic  intuition is that the object category and bounding box should be interrelated, when judging if the insertion is appropriate. By adopting  Bayes' theorem, we arrive at: 
\begin{enumerate}
\item \textbf{Object Recommendation}: 

\begin{equation} P(C|I) = \sum_{B\in \mathcal{B}_I} P(B, C| I). \label{eq:a} \end{equation}

\item \textbf{Scene Retrieval}: \begin{equation} P(I|C) = \frac{P(C|I)P(I)}{P(C)} \propto P(C|I) = \sum_{B\in \mathcal{B}_I} P(B,C|I), \label{eq:b} \end{equation} 

where we perform the maximum a posteriori (MAP) estimation and assume a uniform prior for $P(I)$. 
\item \textbf{Bounding Box Prediction}:

\begin{equation} P(B|C, I) = \frac{P(B,C|I)}{P(C|I)} \propto P(B,C|I), \label{eq:c}\end{equation}

where we rank all bounding boxes $B \in \mathcal{B}_I$ for each given pair of $<C, I>$.
\end{enumerate}

In summary, to achieve our goal (which breaks down to three subtasks), we need an algorithm to estimate $P(B,C|I)$, which is discussed in the next subsection.

\subsection{Modeling the Joint Probability Distribution}

\subsubsection{Model Formulation} 
For each image $I$, we obtain a set of $N$ bounding boxes $\mathbb{B}_I = \{B_i\}_{i=1}^N$ for \textit{existing} objects, which is typically the output of a region proposal network (RPN) \cite{DBLP:journals/corr/RenHG015}. 

Note that the candidate bounding box $B$ and category $C$ are conditionally independent with $\mathbb{B}_I$ given $I$, because  $\mathbb{B}_I$ is derivable from $I$. We then model the joint probability distribution $P(B, C|I)$ as follows:

\begin{equation} P(B,C|I) = P(B, C| \mathbb{B}_I, I)  = \sum_{B_i \in \mathbb{B}_I} P(B, C|B_i, I) \label{eq:d} \end{equation}

We represent each context object with a probability distribution over all possible categories. Denoting the set of all categories considered in the context as $\mathbb{C}$, we have

\begin{equation} 
    \begin{aligned} & P(B, C| B_i, I)  
    \\ &= \sum_{C_j \in \mathbb{C}} P(B, C| B_i, C_j, I) P(C_j | B_i, I) 
    \\ &= \sum_{C_j\in \mathbb{C}} P(B, C| B_i, C_j, I) P(C_j | B_j, I) 
    \\&= \sum_{C_j \in \mathbb{C}} P(C| B_i, C_j, I) P(B| C, B_i, C_j, I) P(C_j | B_i, I) 
    \label{eq:e} 
    \end{aligned} 
\end{equation}

where, the last term in the right-hand-side is the output distribution obtained from an object detector \cite{DBLP:journals/corr/RenHG015, DBLP:journals/corr/AndersonHBTJGZ17}. The first term is decided by the co-occurrence frequency of the inserted object $C$ with an localized existing object $(B_i, C_j)$. For simplicity, we drop $B_i$ and approximate this term with $P(C|C_j) = \frac{count(C, C_j)}{count(C_j)}$. The basic intuition is that $B_i$ does not contribute significantly to the \textit{ranking} between categories. For instance, compared to a mouse, a cake is more likely to co-occur with a plate, no matter where the plate is. The second term is an object-level context \cite{rabinovich2007}  term that will  be modeled with a Gaussian mixture model (GMM), as described next.

\subsubsection{Context Modeling with GMM} 
We now focus on the context term in equation \ref{eq:e} that remains unsolved.
Consider the case when $C = \text{clock}$ and $C_j = \text{wall}$. The term $P(B|C, B_i, C_j, I)$ answers the question ``Having observed a wall in a certain place, \textit{where} should we insert a clock?''. Given such a question, a human agent would first identify that a clock is likely to be \textit{mounted on} the wall, then conclude  that the clock is likely to appear in the upper region of the wall, and its size should be much smaller than the wall. Our GMM model simulates the above process to judge each candidate bounding box .

Based on this intuition, we further exploit inter-object relationships as proposed by previous works on scene graphs \cite{xu2017, DBLP:conf/cvpr/JohnsonKSLSBL15}. Denoting the set of all considered relations as $\mathbb{R}$, we get:

\begin{equation} \begin{aligned} & P(B|C, B_i, C_j, I) \\ &= \sum_{r \in \mathbb{R}} P(r| C, B_i, C_j, I) P(B| r, C, B_i, C_j, I) \\&= \sum_{r \in \mathbb{R}} \frac{count(C, r, C_j)}{count(C, C_j)} P(B| r, C, B_i, C_j, I) \end{aligned} \label{eq:f} \end{equation}

Following \cite{DBLP:conf/cvpr/JohnsonKSLSBL15}, we extract pairwise bounding box feature, which encodes the relative position and scale of the inserted object and a context object:

\begin{equation} f(B_1, B_2) = [\frac{x_1 - x_2}{w_2}, \frac{y_1 - y_2}{h_2}, \frac{w_1}{w_2}, \frac{h_1}{h_2}] \end{equation}

where $(x_1, y_1), (x_2, y_2)$ are the bottom-left corners of the 2 boxes, and $(w_1, w_2), (h_1, h_2)$ are the widths and heights respectively. We then train a Gaussian mixture model (GMM) for each annotated $(\text{subject}, \text{relation}, \text{object})$ triple from the Visual Genome \cite{DBLP:journals/corr/KrishnaZGJHKCKL16} dataset:

\begin{equation} \begin{aligned} 
	& \quad P(B| r, C, B_i,  C_j, I) \propto P(B, B_i| r, C, C_j, I) \\
	& \stackrel{t=(C, r, C_j)}{=} \mathbf{gmm}^{(t)}(f(B, B_i)) \\
	&= \sum_{k=1}^K a_k^{(t)} \mathcal{N}( f(B, B_i) | m_k^{(t)}, M_k^{(t)}) \\
\end{aligned} \label{eq:g} \end{equation}

where $\mathbf{gmm}^{(t)}$ denotes the GMM model corresponding to triple $t=(C, r, C_j)$. $K$ is the number of components same for each GMM, which we empirically set to 4 in our experiments. $\mathcal N$ is the normal distribution. $a_k^{(t)}, m_k^{(t)}, M_k^{(t)}$ are the prior, mean, and covariance for the $k$ th component of $\mathbf{gmm}^{(t)}$, which we learn using the EM algorithm implemented by Scikit-learn \cite{scikit-learn}. 
 
 
\subsubsection{Final Model} 
Putting everything together, we have:
\begin{equation} \begin{aligned} & P(B, C | I) \\&\stackrel{Eq. \ref{eq:d}}{=} \sum_{B_i} P(B, C | B_i, I) 
\\&\stackrel{Eq. \ref{eq:e}}{=} \sum_{B_i \in \mathbb{B}_I}\sum_{C_j \in \mathbb{C}} P(B, C | B_i, C_j, I)P(C_j | B_i, I) 
\\&= \sum_{B_i \in \mathbb{B}_I}\sum_{C_j \in \mathbb{C}} P(C | B_i, C_j, I) P(B | C, B_i, C_j, I) P(C_j | B_i, I)
\\&\stackrel{Eq. \ref{eq:f}}{=} \sum_{B_i \in \mathbb{B}_I}\sum_{C_j \in \mathbb{C}} \frac{count(C, C_j)}{count(C_j)} [\sum_{r\in \mathbb{R}} P(r | C, B_i, C_j, I) 
    \\& \quad \quad \quad \quad \quad \quad \quad \quad \quad \quad  P(B| r, C, B_i, C_j, I)] P(C_j | B_i, I)
\\&\stackrel{Eq. \ref{eq:g}}{=} \sum_{B_i \in \mathbb{B}_I}\sum_{C_j \in \mathbb{C}} \frac{count(C, C_j)}{count(C_j)} [\sum_{r\in \mathbb{R}} \frac{count(C, C_j, r)}{count(C, C_j)} 
    \\& \quad \quad \quad \quad \quad \quad \quad \quad \quad \quad  \mathbf{gmm}^{(C, r, C_j)}(f(B, B_i))] P(C_j | B_i, I)
\\&= \sum_{B_i \in \mathbb{B}_I}\sum_{C_j \in \mathbb{C}} \sum_{r\in \mathbb{R}} \frac{count(C, C_j, r)}{count(C_j)}  
	\\& \quad \quad \quad \quad \quad \quad \quad \quad \quad \quad  \mathbf{gmm}^{(C, r, C_j)}(f(B, B_i)) \underbrace{P(C_j | B_i, I)}_\text{object detection}\end{aligned}\end{equation}

\subsection{Implementation Details} \label{subsec:detail}
We adopt the pretrained Faster R-CNN released by \cite{DBLP:journals/corr/AndersonHBTJGZ17} as object detector. We use 10 object categories for insertion (detailed in Sect. \ref{sec:dataset}) and keep the top 20 object categories and top 10 relations from the Visual Genome \cite{DBLP:journals/corr/KrishnaZGJHKCKL16} dataset, sorted by the co-occurrence count with the 10 insertable categories. We consider at most $N = 20$ existing objects with detection threshold of 0.4 for context modeling. For each image $I$ with size $H_I \times W_I$, we sample the candidate bounding boxes $\mathcal{B}_I$ in a sliding window fashion, with window size $w \in \{\frac{1}{8}, \frac{1}{16}\} \max\{H_I, W_I\}$ and stride $s = \frac{1}{2} w$, which generates around 800 candidate boxes per image. We further refine the size of the best ranked box by searching over sizes within interval $[0, \frac{1}{8}] \max\{H_I, W_I\}$ equally discretized into 32 values. A complete, single thread, pure Python implementation on an Intel i7-5930K 3.50GHz CPU and a single Titan X Pascal GPU takes around 4 seconds per image.

\section{Dataset} \label{sec:dataset}
\subsection{Scenes and Objects}
We establish a  test set that consists of fifty scenes from the Visual Genome \cite{DBLP:journals/corr/KrishnaZGJHKCKL16} dataset. The test scenes come from 4 indoor scene types: \textit{living room, dining room, kitchen, office}. The statistic for each scene type is shown in Table \ref{table:stats-st}. 

\begin{table}[!h] \centering
    \begin{tabular}{c c c c}
    \toprule
   \textbf{living room} & \textbf{dining room} & \textbf{kitchen} &  \textbf{office} \\
    \midrule
   15 & 13 & 10 & 12 \\
    \bottomrule
    \end{tabular}
    \caption{Statistics on different scene types}
    \label{table:stats-st}
\end{table}

There are ten insertable objects considered in this experiment, as shown in Table \ref{table:objects}. The same illustrations and specifications are emphasized to the annotators as a standard to ensure consistency for the same category. We choose these insertable objects based on the following principles:
\begin{enumerate}
\item \textbf{Environment}:  Mostly appears indoor;
\item \textbf{Frequency}: Is within the top 150 frequent categories \cite{xu2017} in Visual Genome;
\item \textbf{Flexibility}: Is not generally embedded (e.g. \textit{sink})  or large and clumsy (e.g. \textit{bed}), so that it can be flexibly inserted into a scene;
\item \textbf{Diversity}: Does not have a significant context overlap with other object categories (e.g. \textit{bottle} is not included because we already have \textit{cup}). 
\end{enumerate}

\begin{table} \centering
    \begin{tabular}{m{1cm} m{1.5cm} m{4.5cm}}
    \toprule
   \textbf{Category} & \textbf{Illustration} & \textbf{Specification} \\
    \midrule
    cup & \includegraphics[width=0.2in]{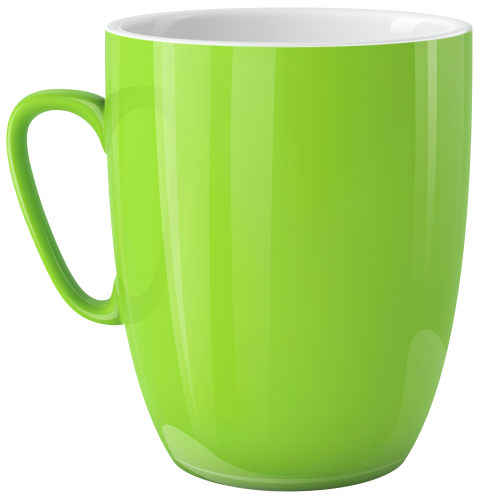} & A cup for drinking water that is medium in size. \\
    \midrule
    spoon & \includegraphics[width=0.2in]{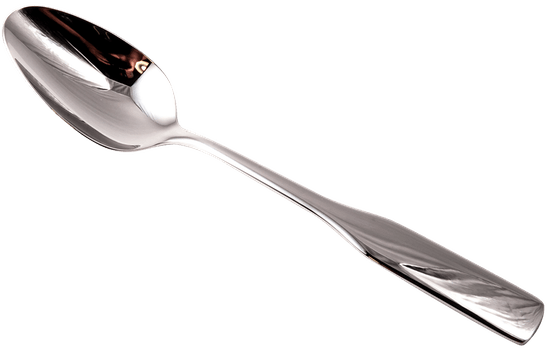} & No specification. \\
    \midrule
    apple & \includegraphics[width=0.2in]{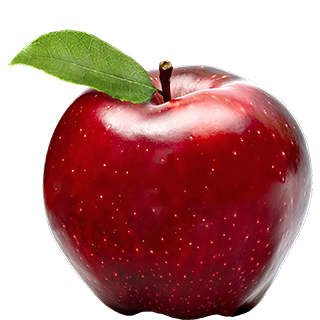} & No specification. \\
    \midrule
    cake & \includegraphics[width=0.2in]{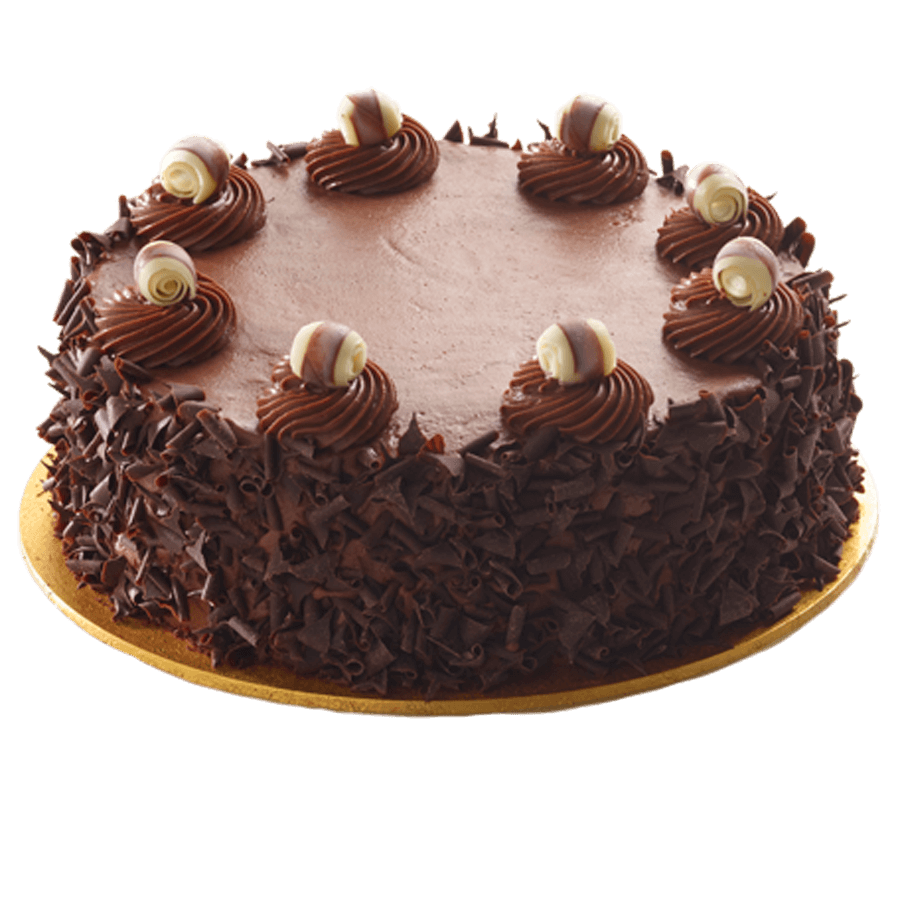} & A small dessert cake (not a big birthday cake). \\
    \midrule
    laptop & \includegraphics[width=0.2in]{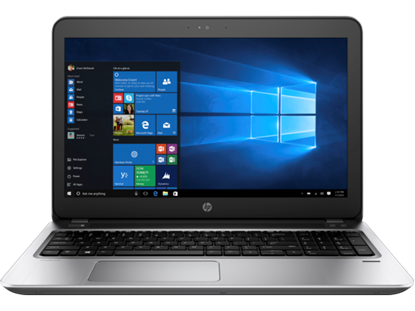} & An open laptop. \\
    \midrule
    mouse & \includegraphics[width=0.2in]{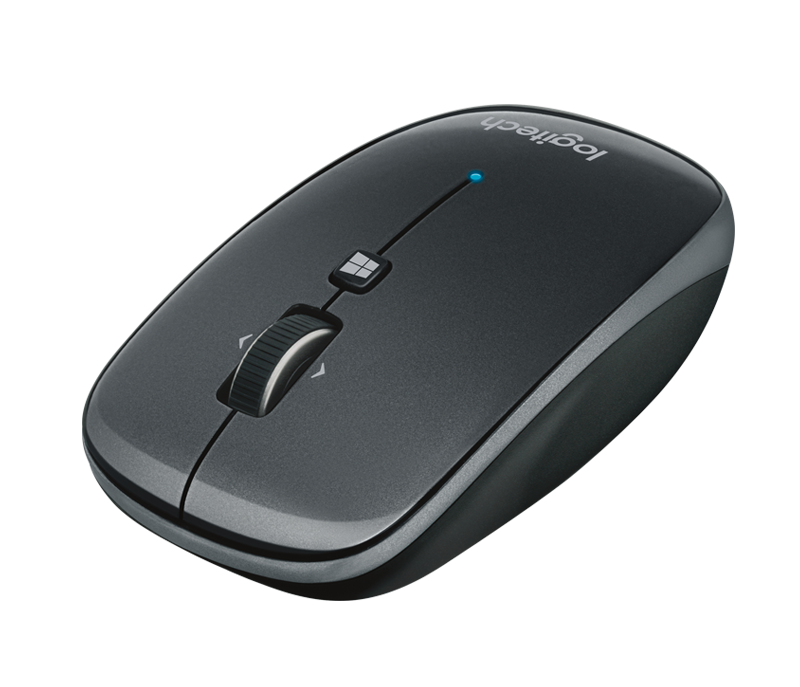} & No specification. \\
    \midrule
    TV & \includegraphics[width=0.2in]{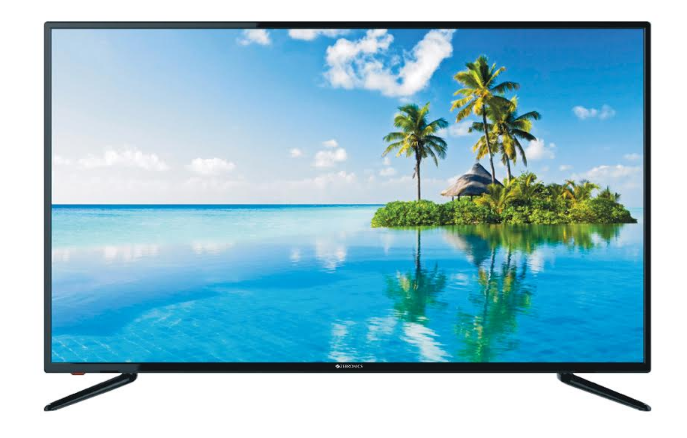} & A LCD TV. \\
    \midrule
    clock & \includegraphics[width=0.2in]{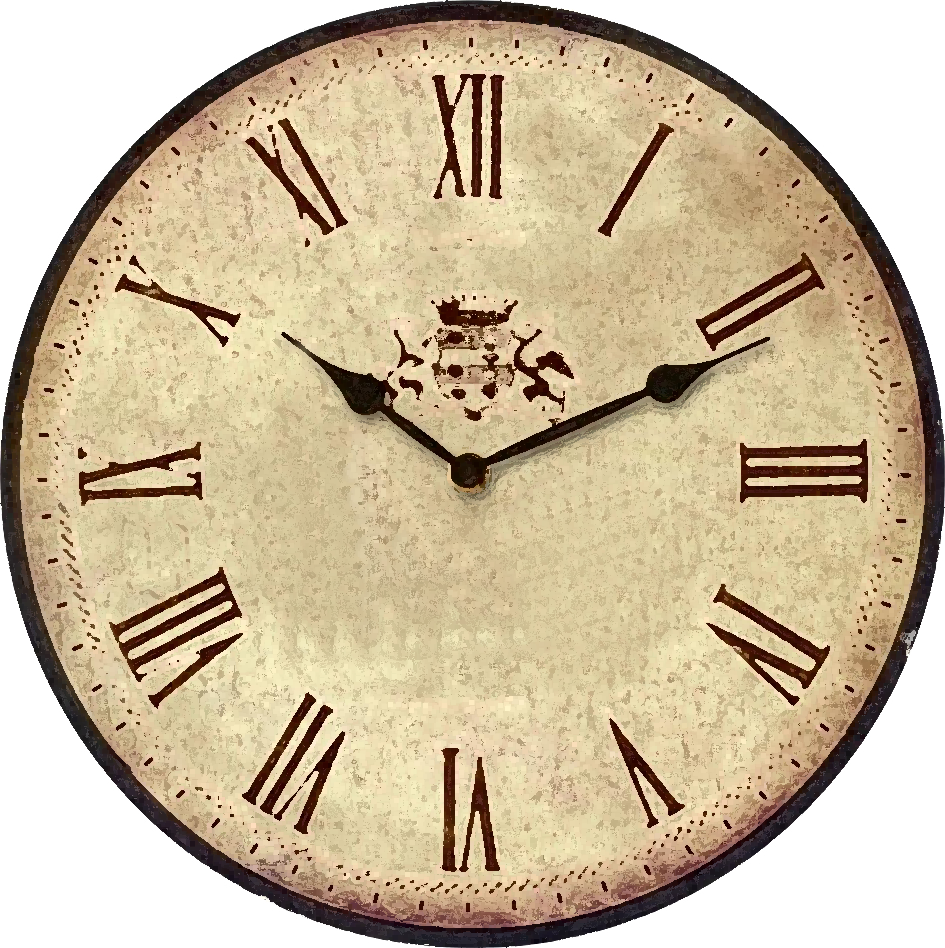} & A normal clock at home (not a watch / alarm clock / bracket clock). \\
    \midrule
    book & \includegraphics[width=0.2in]{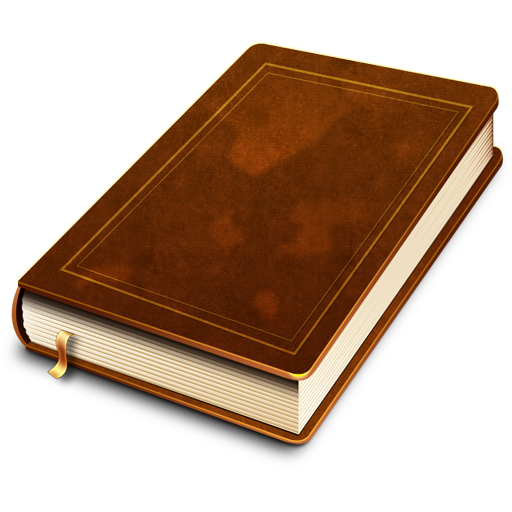} & A closed book that is roughly of B5 size and 200-300 pages. \\
    \midrule
    pillow & \includegraphics[width=0.2in]{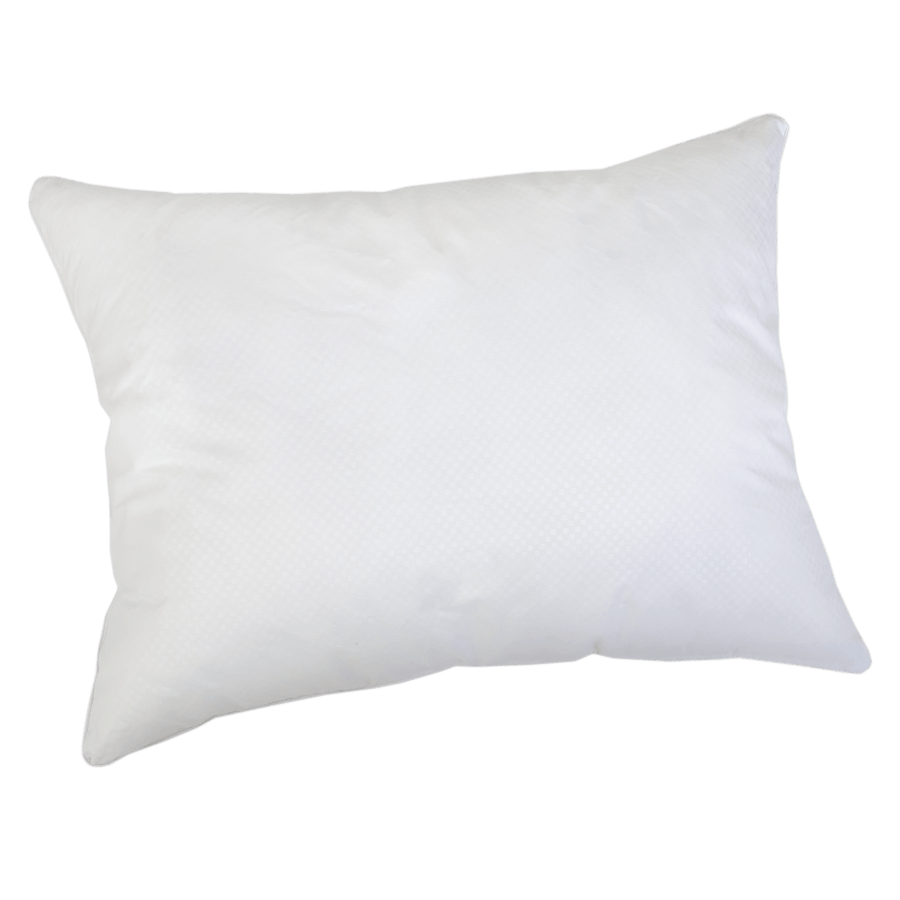} & A rectangle pillow that is commonly placed on sofas, chairs, etc. \\
    \bottomrule
    \end{tabular}
    \caption{Insertable object categories considered in this experiment}
    \label{table:objects}
\end{table}

\subsection{Annotation Guideline}

On average, there are 11 human annotators for each scene. For each scene, the annotator is asked to generate the following annotations: 

\subsubsection{Insertable Categories} 
For each scene, the annotator is encouraged to annotate as much as possible, yet no more than 5 insertable object categories (chosen from the categories in Table \ref{table:objects}). 

\subsubsection{User Preference} 
For each annotated object category, the annotator should assign a preference score ranging from $\{1, 2\}$. The annotators are shown a wide range of different example scenes in advance to ensure that they have consistent criterion towards this preference.
\begin{itemize}
\item \textbf{Score 2 (very suitable)}: Indicates ``this category is very suitable to be inserted into the scene'';
\item \textbf{Score 1 (generally suitable)}: Indicates ``this category can be inserted into this scene, yet not very suitable''.
\end{itemize}

\subsubsection{Bounding Box Size}
For each annotated object category, the annotator should draw a rectangle bounding box, whose longer side equates to the longer side of an appropriate bounding box of the object. We only need 1 freedom for size evaluation because the aspect ratio of the inserted object is typically fixed. 

\subsubsection{Insertable Region} 
\begin{figure}[!h]
    \centering
	\begin{subfigure}[t]{0.3\linewidth}
        \centering\includegraphics[height=1.2in]{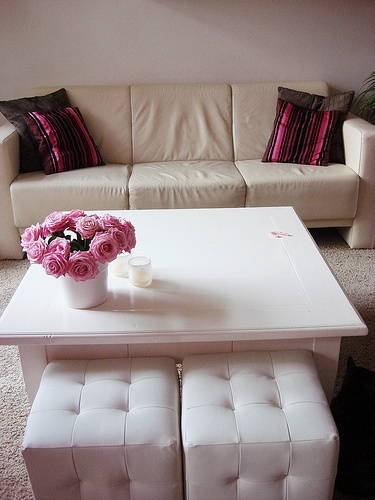}
        \caption{Original image}
        \label{fig:original_image}
    \end{subfigure}
    \begin{subfigure}[t]{0.3\linewidth}
        \centering\includegraphics[height=1.2in]{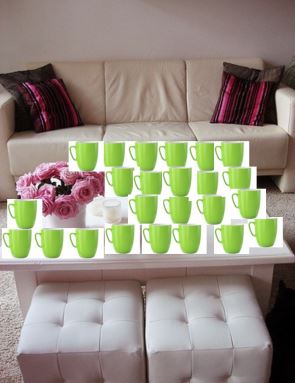}
        \caption{Imagination}
        \label{fig:insertable_region_principle}
    \end{subfigure}
    \begin{subfigure}[t]{0.3\linewidth}
        \centering\includegraphics[height=1.2in]{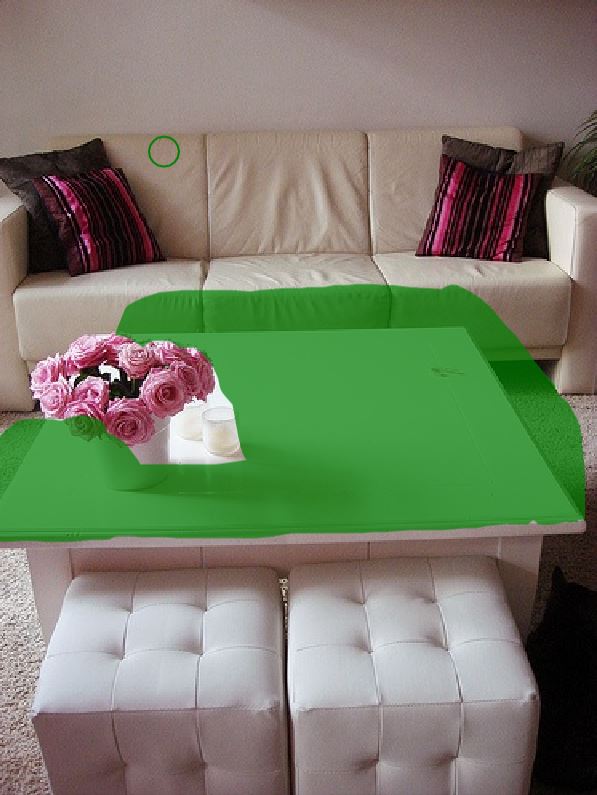}
        \caption{Insertable region}
        \label{fig:insertable_region_final}
    \end{subfigure}
    \caption{Method for drawing the insertable region. This is also the same illustration that we presented to the annotators. Note that in this case, because the cup has a non-zero height, some pixels above the table can also be covered.}
    \label{fig:insertable_region}
\end{figure}

For each annotated object category, the annotator should draw a region. The method for drawing this region is that: Imagine you are holding the object for insertion, and you drag it over all the places that it can be inserted. In this process, the region that can be \textbf{covered} by the object is defined as the \textit{insertable region}, which should be drawn using a brush tool (Fig. \ref{fig:insertable_region}).
 
Note that, different annotators may have different opinions towards this region. For instance, for the scene in Fig. \ref{fig:insertable_region},  some annotators may not include the left-bottom corner of the table when drawing the insertable region. This subjectivity is explicitly allowed within the range of quality control. 

\section{Experiments} \label{sec:exp}

Table \ref{table:show-or} and \ref{table:show-sr} shows qualitative results for \textbf{object recommendation} and \textbf{scene retrieval}, both enhanced by \textbf{bounding box prediction}.  We further quantitatively evaluate our method against existing baselines on our new test set. Task-specific metrics are designed for comprehensive  evaluation. 

We design  experiments for the 3 subtasks systematically. First, for both the \textbf{object recommendation} and \textbf{scene retrieval} subtasks, we compare our system against a statistical baseline, bag-of-categories (BOC), which is  based on category co-occurrence. Second, we separately evaluate the size and location for the \textbf{bounding box prediction} subtask, and compare our results against a recently proposed neural model for person composition \cite{DBLP:journals/corr/TanBCOB17}.  Finally, we report comparisons on both quantitative and qualitative results, which helps interpretations for what is learned by our algorithm.

\begin{table} \centering
    \begin{tabular}{c c c c}
    \toprule
    \textbf{model} & \textbf{nDCG@1} & \textbf{nDCG@3} & \textbf{nDCG@5} \\
    \midrule
    BOC & 43.28\% & 47.81\% & 55.72\% \\
    ours & \textbf{59.06}\%  & \textbf{55.30}\% & \textbf{61.78}\% \\
    \bottomrule
    \end{tabular}
    \caption{Quantitative evaluation for object recommendation}
    \label{table:or}
\end{table}

\begin{table*}
    \centering
    \begin{tabular}{c c c c}
         \includegraphics[width=1.45in]{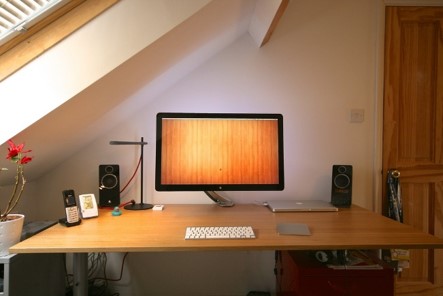} & \includegraphics[width=1.45in]{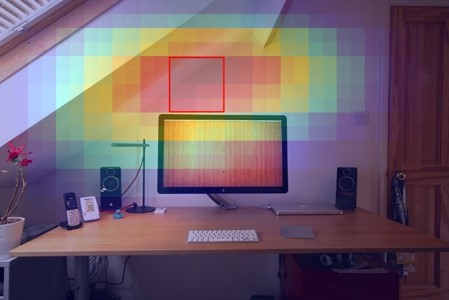} & \includegraphics[width=1.45in]{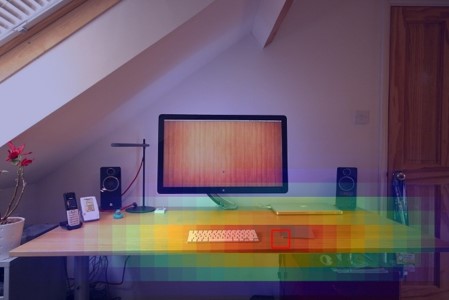} & \includegraphics[width=1.45in]{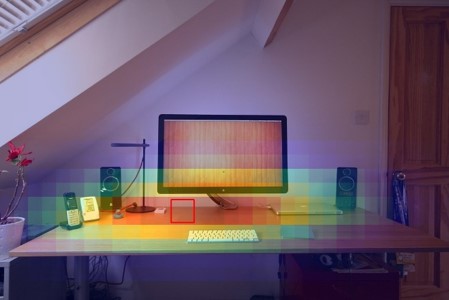} \\
         & \includegraphics[width=1.45in]{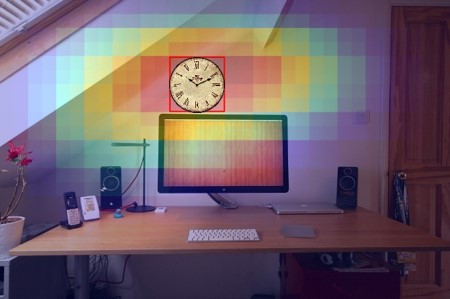} & \includegraphics[width=1.45in]{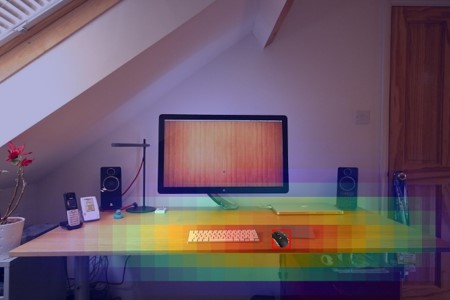} & \includegraphics[width=1.45in]{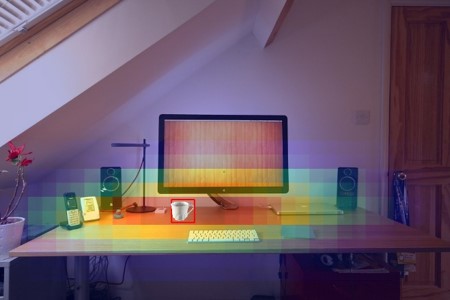} \\
         & \textit{1) clock} & \textit{2) mouse} & \textit{3) cup} \\
         
         \includegraphics[width=1.45in]{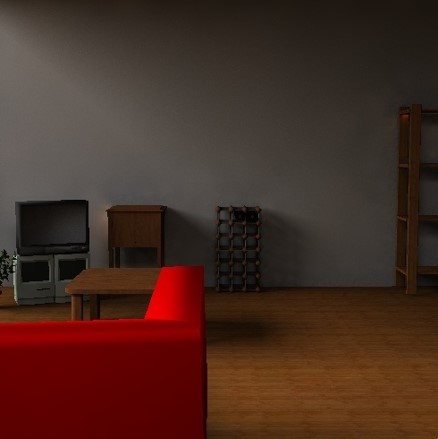} & \includegraphics[width=1.45in]{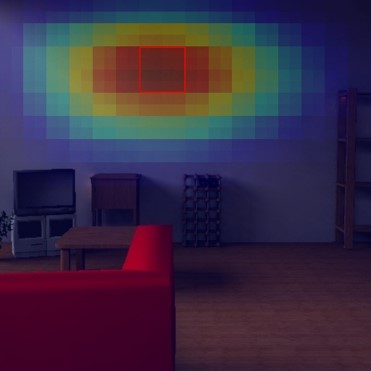} & \includegraphics[width=1.45in]{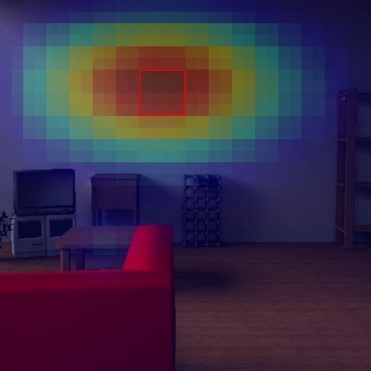} & \includegraphics[width=1.45in]{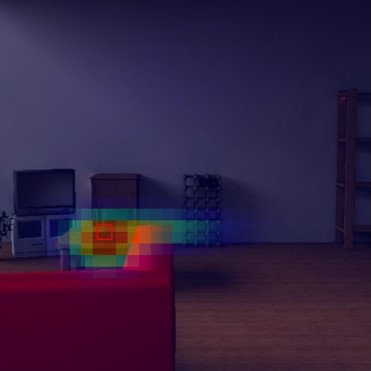} \\
         & \includegraphics[width=1.45in]{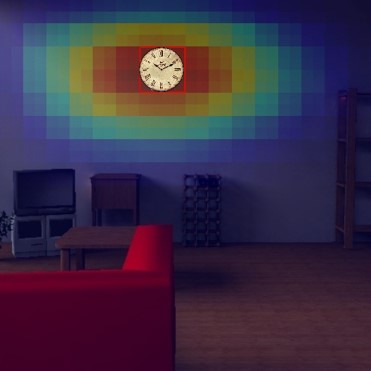} & \includegraphics[width=1.45in]{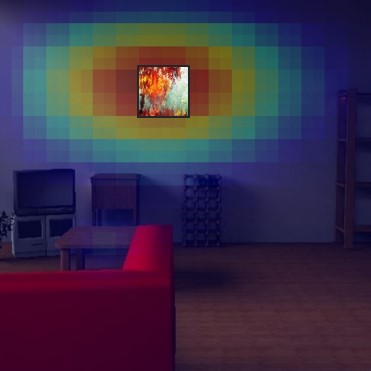} & \includegraphics[width=1.45in]{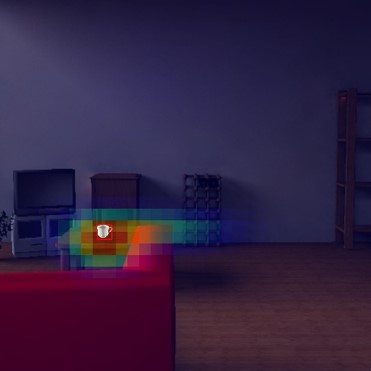} \\
         & \textit{1) clock} & \textit{2) tv} & \textit{3) cup} \\
         
         \includegraphics[width=1.45in]{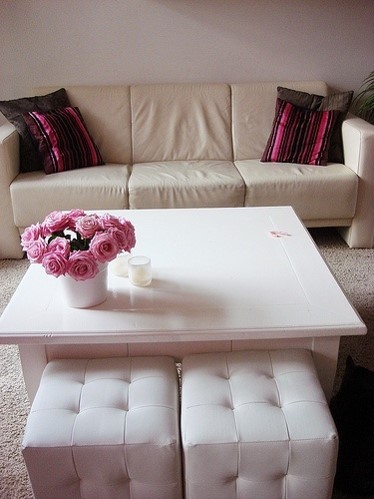} & \includegraphics[width=1.45in]{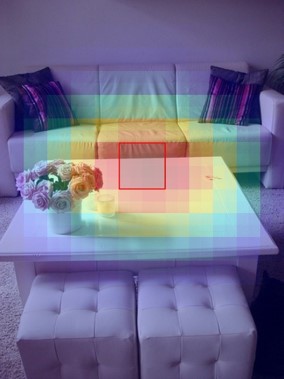} & \includegraphics[width=1.45in]{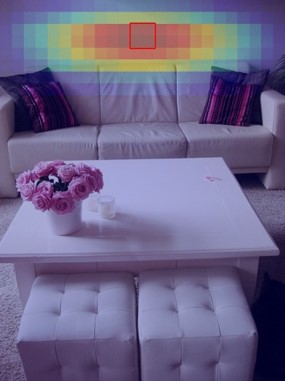} & \includegraphics[width=1.45in]{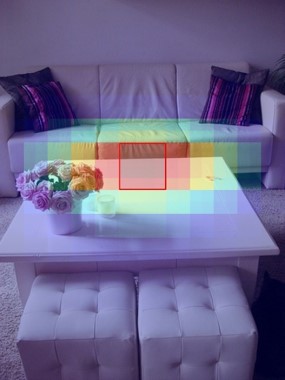} \\
         & \includegraphics[width=1.45in]{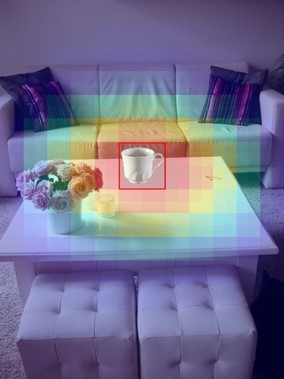} & \includegraphics[width=1.45in]{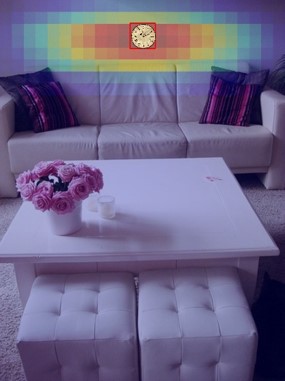} & \includegraphics[width=1.45in]{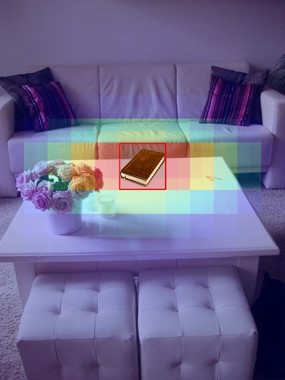} \\
         & \textit{1) cup} & \textit{2) clock} & \textit{3) book} \\
    \end{tabular}
    \caption{Object recommendation results. The first row for each demo is the recommended bounding box (\ref{subsubsec:bestbox}) and indicative heatmap (\ref{subsubsec:hm}) \textit{automatically} generated for the top 3 recommended object categories (\ref{subsec:objrec}). The second row is generated using \textit{manually} selected yet \textit{automatically} pasted object segments (for illustration purpose only).}
    \label{table:show-or}
\end{table*}

\begin{table*}
    \centering
    \begin{tabular}{c}
         \begin{subfigure}[t]{\linewidth} \centering \includegraphics[width=6in]{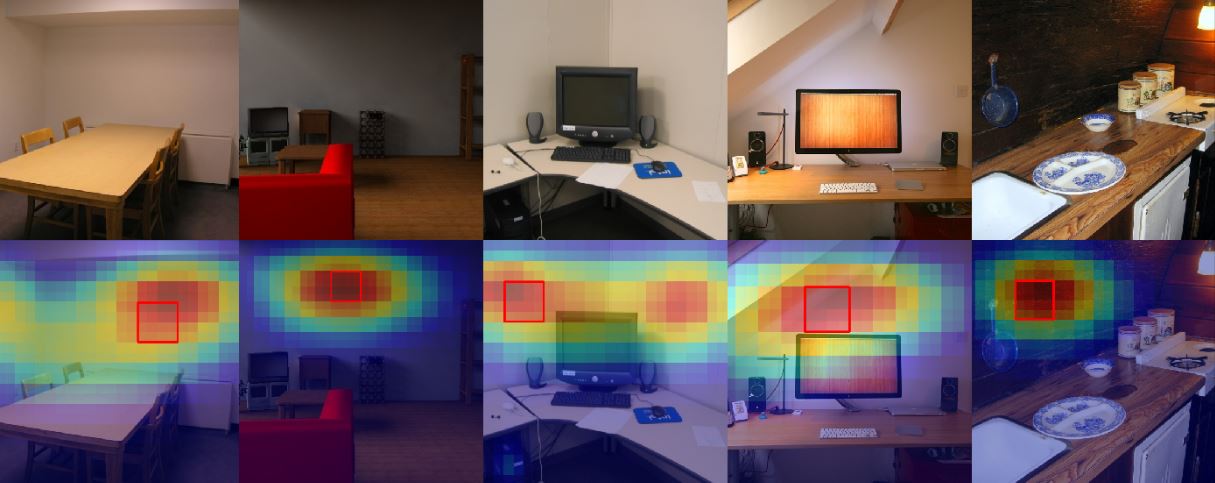} \caption{\textit{clock}} \end{subfigure} \\
         \\
         \begin{subfigure}[t]{\linewidth} \centering \includegraphics[width=6in]{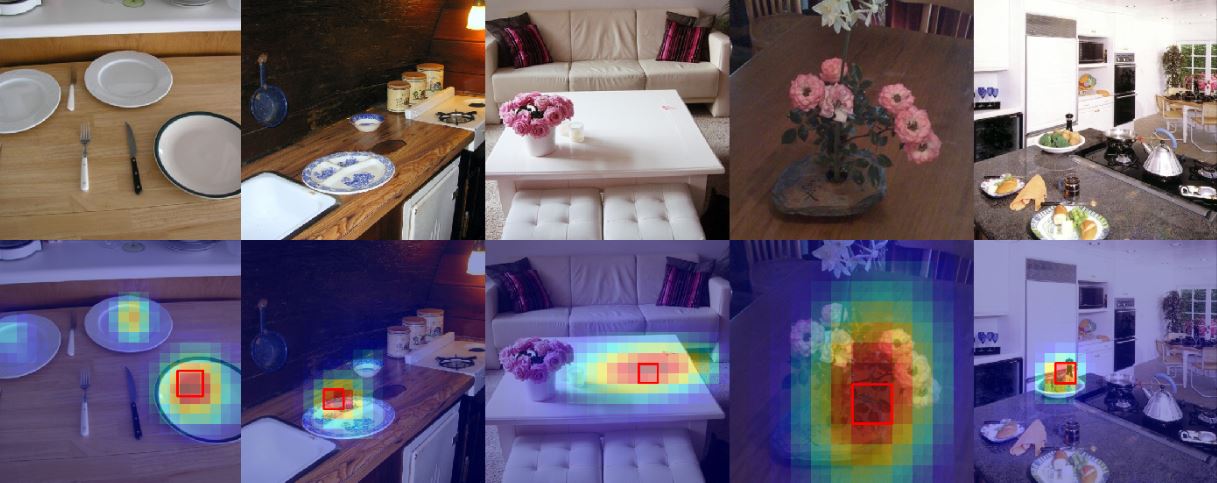} \caption{\textit{apple}} \end{subfigure} \\
         \\
         \begin{subfigure}[t]{\linewidth} \centering \includegraphics[width=6in]{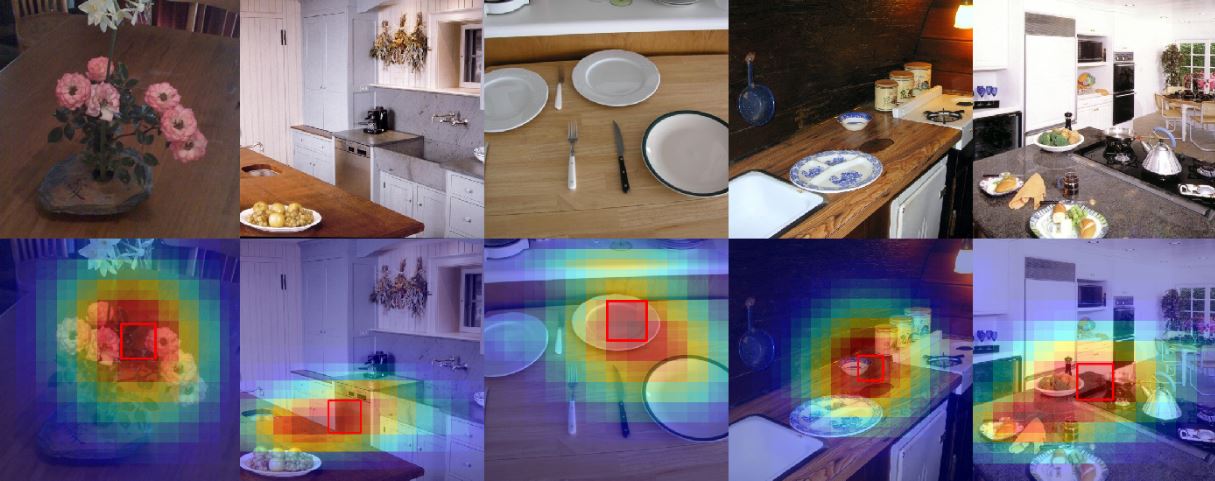} \caption{\textit{cup}}  \end{subfigure}\\
    \end{tabular}
    \caption{Scene retrieval results: top 5 retrieved scenes (\ref{subsec:senret}) for \textit{clock, apple, cup}. The first row for each category shows the original images, and the second row shows the scenes overlaid with \textit{automatically} generated bounding boxes (\ref{subsubsec:bestbox}) and heatmaps (\ref{subsubsec:hm}). }
    \label{table:show-sr}
\end{table*}

\subsection{Object Recommendation} \label{subsec:objrec}

We adopt the normalized discounted cumulative gain (nDCG) \cite{Jarvelin:2002:CGE:582415.582418}, which is an indicator widely used in information retrieval (IR) for ranking quality. We use this as the metric for object recommendation. Because the desired output for this subtask is a ranked list, and each item is annotated with a gain reflecting user preference, nDCG is a perfect choice for evaluation.

\subsubsection{Metric Formulation}
For $n$ images and $m_i$ annotators for the $i$ th image, the averaged nDCG@K is defined as

\begin{equation} \overline{nDCG@K} = \frac{1}{n} \sum_{i=1}^n \frac{1}{m_i} \sum_{j=1}^{m_i} nDCG@K(i, j) \end{equation}
where $nDCG@K(i, j)$ measures the ranking quality for the top-K recommended object categories  of the $i$ th image, with regard to the ground truth user preference scores provided by the $j$ th annotator. 

\subsubsection{Quantitative Results} The baseline method, bag-of-categories (BOC), regards each image as a bag of existing objects, and ranks all candidate objects by the sum of co-occurrences with the existing objects. BOC borrows idea from the simple yet effective bag-of-words (BOW) model \cite{wikibow} in natural language processing, which ignores the structural information and only keeps the statistical count. 

The quantitative comparison between our system and BOC is shown in Table \ref{table:or}. We evaluate nDCG at the top-1, top-3, top-5 results respectively, because there are at most 5 annotations per image. As demonstrated by the results, our method achieves consistent improvements as compared to BOC.

\subsubsection{Qualitative Analysis} 

\begin{figure}[!h]
    \centering
	\begin{subfigure}[t]{0.3\linewidth}
        \centering\includegraphics[height=0.8in]{assets/2352548.jpg}
        \caption{ours: \textit{mouse} \\ BOC: \textit{clock}}
        \label{fig:1a}
    \end{subfigure}
    \begin{subfigure}[t]{0.3\linewidth}
        \centering\includegraphics[height=0.8in]{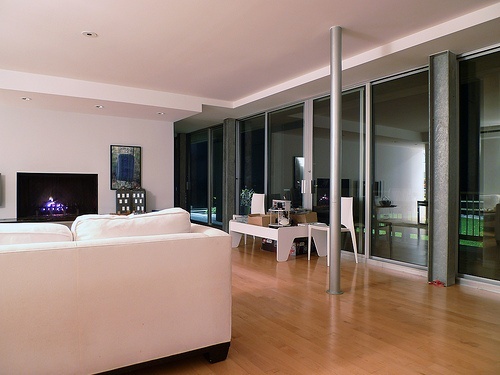}
        \caption{ours: \textit{clock} \\ BOC: \textit{laptop}}
        \label{fig:1b}
    \end{subfigure}
    \begin{subfigure}[t]{0.3\linewidth}
        \centering\includegraphics[height=0.8in]{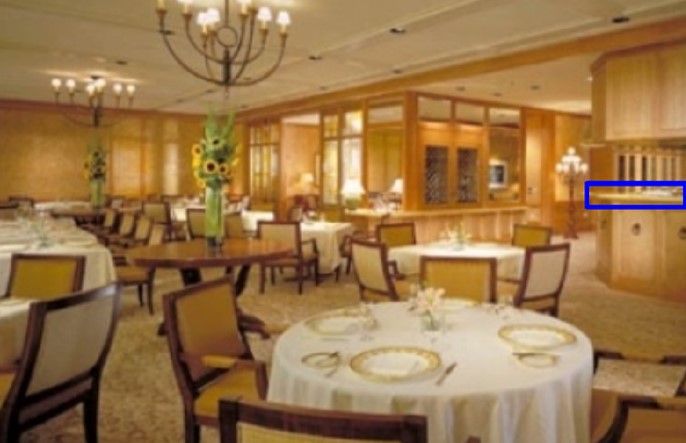}
        \caption{ours: \textit{cup} \\ BOC: \textit{book}}
        \label{fig:1c}
    \end{subfigure}
    \caption{Qualitative comparison on top recommended object}
    \label{fig:toprec}
\end{figure}

The largest gain of our method over baseline is reflected by nDCG@1, i.e. the top result. Fig. \ref{fig:toprec} shows qualitative comparison against BOC on top 1 recommendation.  In Fig. \ref{fig:1a}, the baseline wrongly recommends a clock because there are 2 detected walls. Whereas, our system recognizes that most candidate boxes for clock lead to unreasonable relative positions with the walls. In Fig. \ref{fig:1b}, the baseline recommends a laptop due to high co-occurrence of pair (laptop, table). However, the table in this scene is too small, disabling any noticeable bounding box for insertion. In Fig. \ref{fig:1c},  the baseline recommends a book because there's a mis-detection for a small shelf to the edge of the background (the blue box, which is actually a counter), which is almost ignored by our system for the same reason as in Fig. \ref{fig:1b}.

In summary, the key advantage of our algorithm over baseline is that we not only consider the co-occurrence frequency, but also take into account the relative locations and relationships between the inserted object and context objects. This enables our system to bypass candidate categories with high co-occurrence counts yet unreasonable placements; and to also be more robust when faced with detection failures.

\subsection{Scene Retrieval} \label{subsec:senret}

\begin{figure*}[h]
    \centering
	\begin{subfigure}[t]{0.48\linewidth}
        \centering\includegraphics[width=0.95\linewidth]{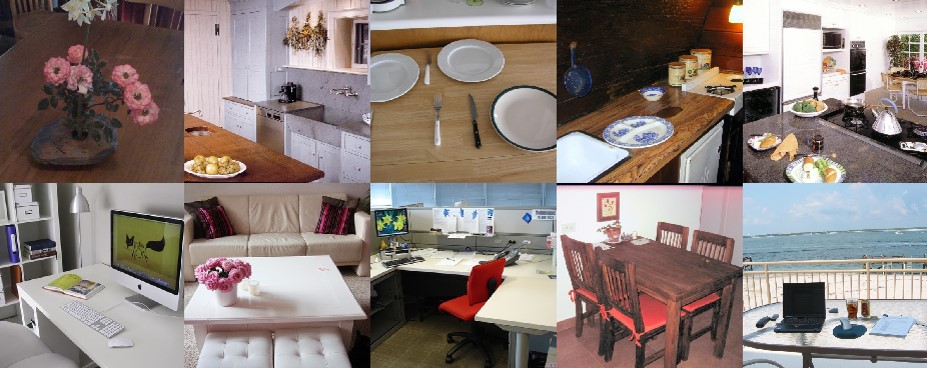}
        \caption{\textit{cup} | ours}
        \label{fig:2a}
    \end{subfigure}
    \begin{subfigure}[t]{0.48\linewidth}
        \centering\includegraphics[width=0.95\linewidth]{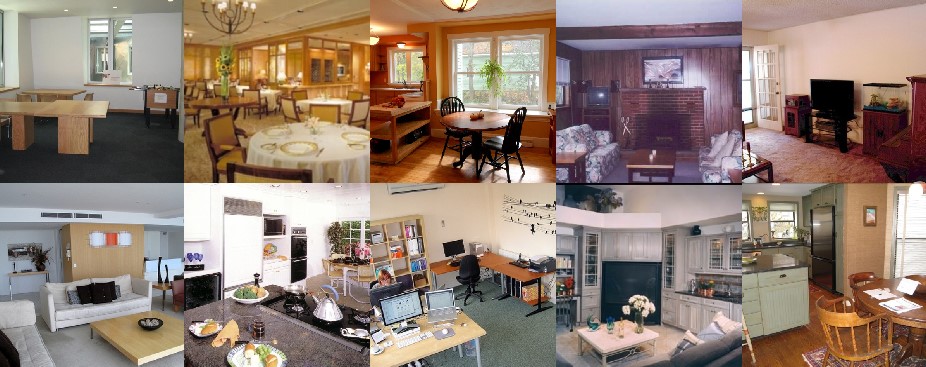}
        \caption{\textit{cup} | BOC}
        \label{fig:2b}
    \end{subfigure}
    \begin{subfigure}[t]{0.48\linewidth}
        \centering\includegraphics[width=0.95\linewidth]{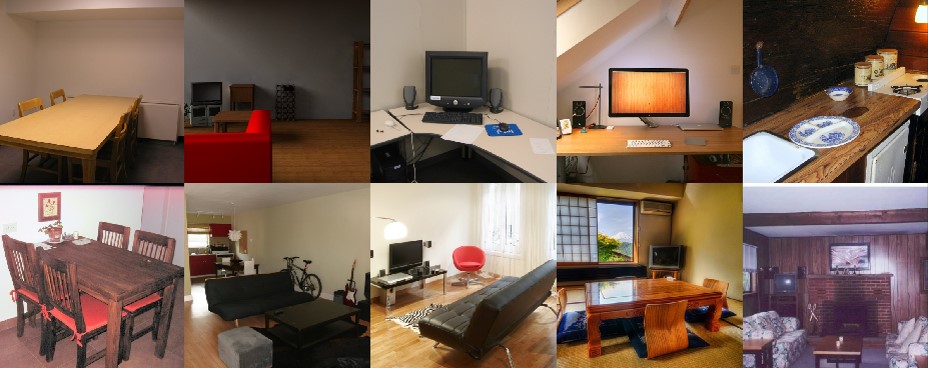}
        \caption{\textit{clock} | ours}
        \label{fig:2c}
    \end{subfigure}
    \begin{subfigure}[t]{0.48\linewidth}
        \centering\includegraphics[width=0.95\linewidth]{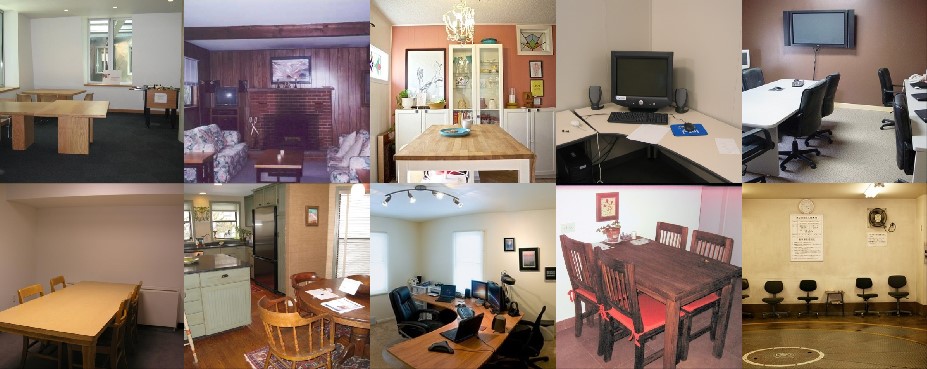}
        \caption{\textit{clock} | BOC}
        \label{fig:2d}
    \end{subfigure}
    \caption{Qualitative comparison on top 10 recommended scenes}
    \label{fig:top10rec}
\end{figure*}

Similarly, for the scene retrieval subtask, we also adopt nDCG as a metric for ranked image list. 

\subsubsection{Metric Formulation}
For $n$ insertable categories and $m$ candidate images for each category, the averaged nDCG@K is defined as 

\begin{equation} \overline{nDCG@K} = \frac{1}{n} \sum_{i=1}^n \frac{1}{m} \sum_{j=1}^m nDCG@K(i, j) \end{equation}
where $nDCG@K(i, j)$ measures the ranking quality of the top-K retrieved scene images for the $i$ th category, with regard to the ground truth user preference scores provided by the $j$ th annotator.

\subsubsection{Quantitative Results} The quantitative comparison between our system and BOC is shown in table \ref{table:sr}.  We evaluate nDCG at $K = 1, 10, 20$ respectively, in consideration of the fact that there are 50 candidate images in total. Again, we outperform the baseline by a remarkable margin. 

\begin{table}[!h] \centering
    \begin{tabular}{c c c c }
    \toprule
    \textbf{model} & \textbf{nDCG@1} & \textbf{nDCG@10} & \textbf{nDCG@20} \\
    \midrule
    BOC & 60.00\% & 50.70\% & 56.03\% \\
    ours & \textbf{65.45}\%  & \textbf{54.87}\% & \textbf{58.25}\% \\
    \bottomrule
    \end{tabular}
    \caption{Quantitative evaluation for scene retrieval}
    \label{table:sr}
\end{table}

\subsubsection{Qualitative Analysis} Fig. \ref{fig:top10rec} shows qualitative comparison against BOC on top 10 retrieved scenes. Intuitively, our system prefers scenes whose supportive objects that are visually large, continuous or close to the user,  while the baseline is typically biased towards scenes with more relevant objects. This is due to the fact that only boxes that lead to reasonable relationships will contribute significantly to $P(B,C|I)$, while BOC is agnostic to the spatial structure of the context objects.

\subsection{Bounding Box Prediction}
We evaluate the size and location of the predicted bounding box separately. The baseline for this subtask is the neural approach proposed by \cite{DBLP:journals/corr/TanBCOB17}. \cite{DBLP:journals/corr/TanBCOB17} builds an automatic two-stage pipeline for inserting a person's segment into an image. It first determines the best bounding box using the dilated convolution networks \cite{DBLP:journals/corr/YuK15}, then retrieves a context-compatible person segment from a database. 

Here, we compare our system's performance on bounding box prediction, against the first stage of \cite{DBLP:journals/corr/TanBCOB17}. We adopt the same object detector  \cite{DBLP:journals/corr/AndersonHBTJGZ17} with the same confidence threshold as in our experiments, and the same training settings for \cite{DBLP:journals/corr/TanBCOB17} as reported in its supplementary material.

For size prediction, we design a single metric to measure the similarity of 2 lengths. For location prediction, however, we design 2 different metrics for automatic use cases and manual use cases, respectively. The automatic use case would require an API that returns the best ranked bounding box, while the manual use case would prefer a heatmap as an intuitive hint. We will discuss these  3 metrics and different use cases in detail.

\subsubsection{Metric Formulation | Size}
For a bounding box $B$ with height $h_B$ and width $w_B$, we define its \textit{box size} $s_B = \max\{h_B, w_B\}$. We then define a metric that evaluates how close is the ground truth box compared with the predicted box, under the measurement of box size. Note that we only preserve 3 freedoms for a box, because the aspect ratio of the inserted object segment should be predetermined. 

Given $n$ images, and $m_i$ annotators for the $i$ th image, for a specific category $C$, we define the average intersection over union (IoU) score for box size as:

\begin{equation} IoU_{size}^{(C)} = \frac{1}{n} \sum_{i=1}^n \frac{\sum_{j=1}^{m_i} \delta_{ij}^{(C)} IoU_{size}({g_{ij}}^{(C)}, {s_{i}}^{(C)})}{\sum_{j=1}^{m_i} \delta_{ij}^{(C)}}   \end{equation}
\begin{equation} IoU_{size}(g, s) = \frac{\min(g, s)}{\max(g, s)} \end{equation}
where, $\delta_{ij}^{(C)} = \begin{cases} 1 & \text{C is annotated by annotator j in image i} \\ 0 & \text{otherwise} \end{cases}$ , $g_{ij}^{(C)}$ is the ground truth box size provided by annotator $j$ in image $i$ for category $C$, $s_i^{(C)}$ is the predicted box size in image $i$ for category $C$. $IoU_{size}(g, s)$ has an upper bound of 1.0 (when $g = s$), and a lower bound of 0.0 (when $g$ and $s$ are drastically different).

\subsubsection{Metric Formulation | Location, Best Box} \label{subsubsec:bestbox}

The best recommended box would be crucial to an automatic application, such as an automatic preview software. Hence, this experiment evaluates whether the best recommended box is in a reasonable location. We consider the location of a bounding box as reasonable, if it is contained within the insertable region annotated by the user. 

\begin{figure}[!h]
    \centering
	\begin{subfigure}[t]{0.3\linewidth}
        \centering\includegraphics[height=1.2in]{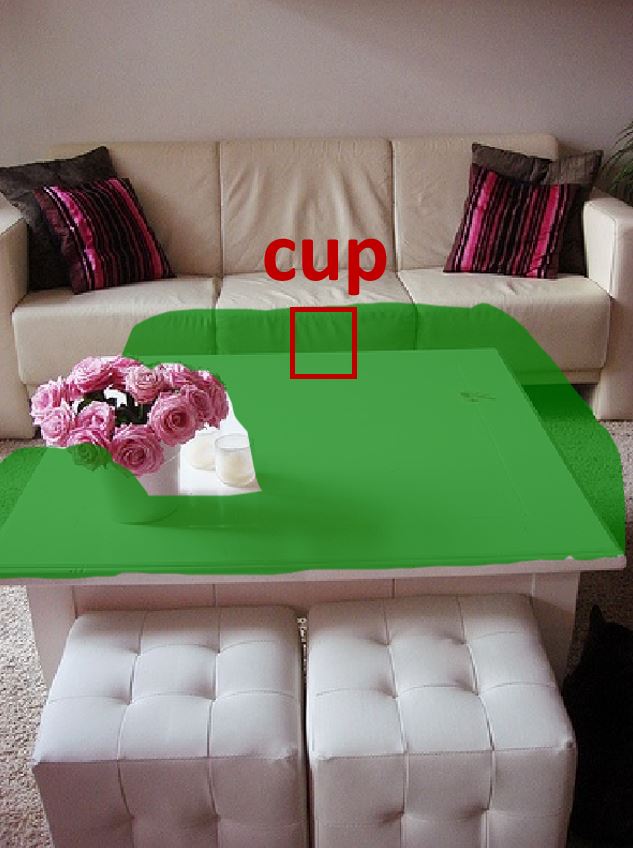}
        \caption{}
        \label{fig:3a}
    \end{subfigure}
    \begin{subfigure}[t]{0.3\linewidth}
        \centering\includegraphics[height=1.2in]{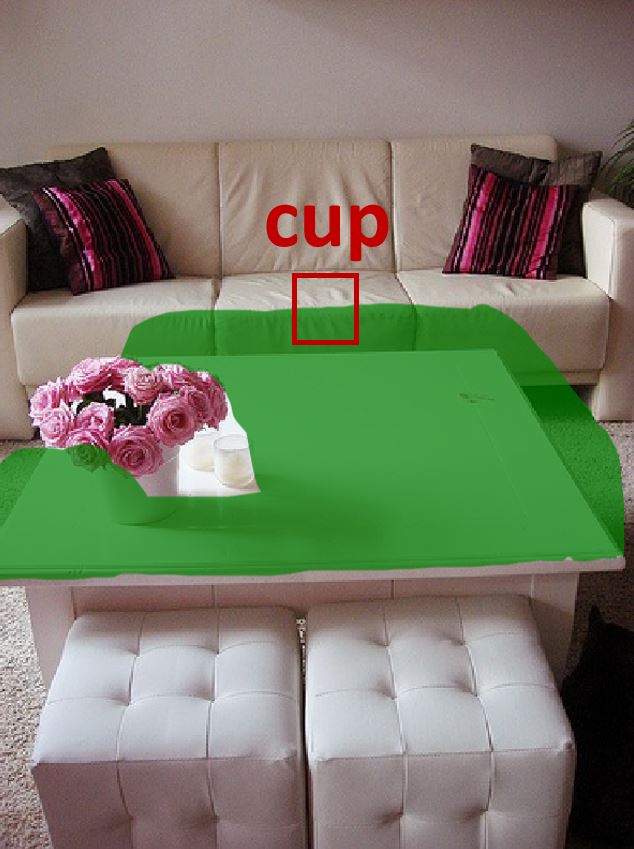}
        \caption{}
        \label{fig:3b}
    \end{subfigure}
    \begin{subfigure}[t]{0.3\linewidth}
        \centering\includegraphics[height=1.2in]{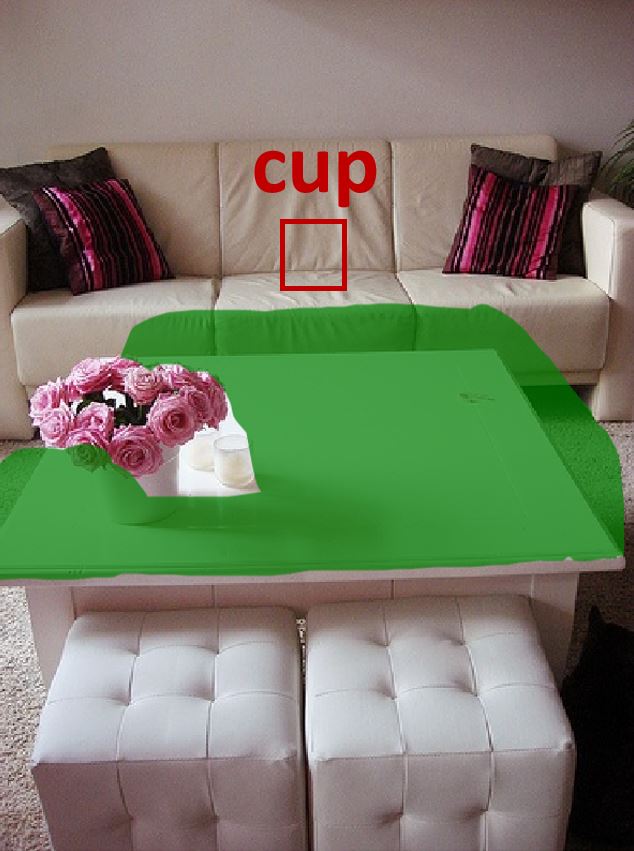}
        \caption{}
        \label{fig:3c}
    \end{subfigure}
    \caption{Metric for best box: A box is regarded as reasonable, if it is contained within the insertable region annotated by the user (Fig. \ref{fig:3a}). A box that slightly exceeds this region (Fig. \ref{fig:3b}) is not good enough, yet still visually better than a box that is an outlier (Fig. \ref{fig:3c}). For best box evaluation, the difference between \textbf{accuracy} and \textbf{strict accuracy} is that for cases like Fig. \ref{fig:3b}, the former one counts the fraction of area that is included in the insertable region, whereas the later one only counts valid boxes as in Fig. \ref{fig:3a}.}
    \label{fig:bestbox}
\end{figure}

Note that this criterion can be biased towards smaller boxes. We address this drawback in 2 aspects:  First, larger boxes that slightly exceeds the insertable region may still have a non-zero contribution to this metric; Second, unreasonably small boxes will pull down the size prediction score accordingly.

Given $n$ images, and $m_i$ annotators for the $i$ th image, for a specific category $C$, we define the average accuracy for the location of best recommended box as

\begin{equation} \text{accuracy}_{loc}^{(C)} = \frac{1}{n} \sum_{i=1}^n \frac{\sum_{j=1}^{m_i} \delta_{ij}^{(C)} \text{accuracy}_{loc}(g_{ij}^{(C)}, B_i^{(C)})}{\sum_{j=1}^{m_i} \delta_{ij}^{(C)}} \label{eq:accloc} \end{equation}
\begin{equation} \text{accuracy}_{loc}(g, B) = \frac{\text{intersection area of g and B}}{\text{area of B}} \end{equation}
where, $\delta_{ij}^{(C)}$ shares the same meaning as before. $g_{ij}^{(C)}$ is the ground truth insertable region drawn by annotator $j$ in image $i$ for category $C$, $B_i^{(C)}$ is the best recommended box in image $i$ for category $C$. $\text{accuracy}_{loc}(g, B)$ has an upper bound of 1.0 (when $B$ is entirely contained within $g$), and a lower bound of 0.0 (when $B$ is entirely outside $g$).

Furthermore, if we only regard bounding boxes that are \textbf{fully} contained by the insertable region as reasonable, we can define a stricter metric by substituting the $\text{accuracy}_{loc}$ in Eq. \ref{eq:accloc} with a binary indicator function $\mathbb{I}(g, B)$ , which is set to 1 if and only if $B$ is fully contained by $g$. We denote this metric as the ``strict accuracy''. This metric excludes boxes that are partially contained by the insertable region, and only counts for valid boxes that are entirely covered.

\subsubsection{Metric Formulation | Location, Heatmap} \label{subsubsec:hm}
This metric evaluates the score distribution of all sampled boxes, which we further convert into an intuitive pixel-level representation. We denote this representation as a \textit{heatmap}.

Specifically, we generate a heatmap by adding the score of each sampled box to all its contained pixels. The heat value at each pixel hence approximates the probability that it is contained within at least one insertable box \footnote{For each pixel $p$ in image $I$ contained within candidate boxes $B_1, B_2, ..., B_k$, for a specific category $C$, we have $$\begin{aligned} & P(p \text{ is contained by at least 1 insertable box}) \\&= 1 - \prod_{i=1}^k (1 - P(B_i|I,C))\approx \sum_{i=1}^k P(B_i|I,C) \end{aligned}$$
when $P(B_i|I,C) << 1$. Typically, there are around 800 candidate boxes per image, therefore the numerical value of $P(B_i|I,C)$ is reasonably small to make this approximation.}.  This representation is compatible, and hence directly comparable, with the \textit{insertable region} provided by the user.

\begin{figure}[!h]
    \centering
    \includegraphics[width=3in]{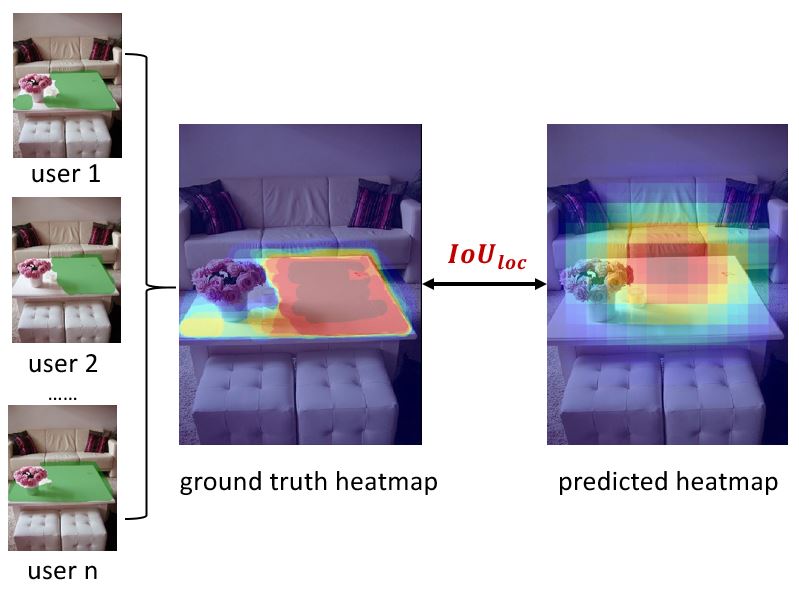}
    \caption{Metric for heatmap: The heat value at each pixel represents the probability that it is contained within at least 1 insertable box. We take the average insertable region over all users as the ground truth heatmap, and test the consistency of this pixel-level probability distribution between ground truth and prediction. This metric measures the system's ability to approximate the hint provided by a human.}
    \label{fig:hmeval}
\end{figure}

Note that the heatmap does \textit{not} support any programmatical usages, but only aims to provide a clear user hint. We do not adopt the distribution of the left-bottom corner or the stand position \cite{DBLP:journals/corr/TanBCOB17} because not all the insertable categories are supported from the bottom (e.g. TV). Hence, a heatmap that dissolves the probability of each box into its inner pixels is more cognitively consistent across different categories. 

Given $n$ images, and $m_i$ annotators for the $i$ th image, for a specific category $C$, we define the average IoU for the heatmap as (illustrated in Fig. \ref{fig:hmeval})

\begin{equation} IoU_{loc}^{(C)} = \frac{1}{n} \sum_{i=1}^n \frac{IoU_{loc}(\sum_{j=1}^{M_i} \delta_{ij}^{(C)} g_{ij}^{(C)} , h_i^{(C)})}{\sum_{j=1}^{M_i} \delta_{ij}^{(C)}} \label{eq:iouheatmap}\end{equation}
\begin{equation} IoU_{loc} (g, h) = \frac{\sum_p \min(g_p, h_p)}{\sum_p \max(g_p, h_p)} \label{eq:heatmap} \end{equation}
In Eq. \ref{eq:iouheatmap}, $\delta_{ij}^{(C)}$ shares the same meaning as before.  $g_{ij}^{(C)}$ is the ground truth insertable region drawn by annotator $j$ in image $i$ for category $C$, $h_i^{(C)}$ is the predicted heatmap for category $C$ in image $i$.

In Eq. \ref{eq:heatmap}, $g$  denotes the averaged ground truth insertable region, and $h$ denotes the predicted heatmap.  $p$ iterates through all pixels of $g$ and $h$. We normalize $g$ and $h$ such that they each sums to 1.0. This definition for IoU has a maximum value 1.0 when $g$ and $h$ are exactly the same, and a minimum value 0.0 when they are absolutely disjoint.

\subsubsection{Quantitative Results} We report the average IoU for size, the average accuracy for best recommended location, and the average IoU for heatmap, over all insertable categories. We refine the location heatmap generated by \cite{DBLP:journals/corr/TanBCOB17} by adding the heat value at each stand position to pixels in the corresponding box to match our heatmap definition. As shown in Table \ref{table:t3}, we achieve consistent improvement over the baseline in all metrics designed for bounding box prediction.

\begin{table}
\centering
\begin{tabular}{c c c c c}
      \toprule
      \textbf{\makecell{model}} & \textbf{\makecell{IoU (size)}} & \textbf{\makecell{accuracy \\ (location, \\best box)}}  & \textbf{\makecell{strict accuracy \\ (location, \\best box)}}  & \textbf{\makecell{IoU \\ (location, \\heatmap)}} \\
      \midrule
      \cite{DBLP:journals/corr/TanBCOB17} & 59.79\% & 12.50\% &  2.87\% & 9.14\% \\
      ours & \textbf{64.90\%} & \textbf{33.65\%} & \textbf{9.46\%} & \textbf{18.23\%}  \\
      \bottomrule
\end{tabular}
\caption{Quantitative evaluation for bounding box prediction}
\label{table:t3}
\end{table}

\subsubsection{Qualitative Analysis} Table \ref{table:bbox_pred} shows the qualitative comparison against \cite{DBLP:journals/corr/TanBCOB17} on bounding box prediction. We outperform baseline significantly, especially on location prediction. Possible reasons include: 1) The baseline employs an impainting model to generate fake background images that do not contain the target object, which leads to error propagation throughout the downstream training process; 2) The Visual Genome \cite{DBLP:journals/corr/KrishnaZGJHKCKL16} dataset is relatively small, and images containing non-human objects (i.e. the insertable categories considered in this paper) are even fewer. We do not use larger datasets such as MS-COCO \cite{coco}, because many important context object categories (e.g. \textit{desk}, \textit{counter}, \textit{wall}, etc.) are not annotated. The data-driven nature of neural network hence limits the performance of \cite{DBLP:journals/corr/TanBCOB17}.

\renewcommand{\tabcolsep}{2pt}

\begin{table*}[h]
    \centering
    \begin{tabular}{>{\centering\arraybackslash} m{0.8cm} >{\centering\arraybackslash} m{1.2in} >{\centering\arraybackslash} m{1.2in} >{\centering\arraybackslash} m{1.2in} >{\centering\arraybackslash} m{1.2in} >{\centering\arraybackslash} m{1.2in}}
         input & \includegraphics[width=1.2in, height=0.8in]{assets/cup1.jpg} & \includegraphics[width=1.2in, height=0.8in]{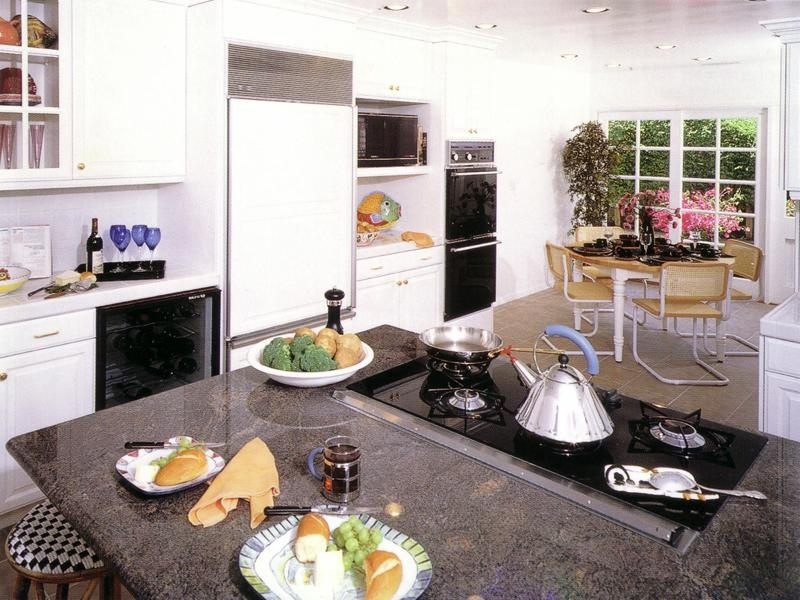} & 
         \includegraphics[width=1.2in, height=0.8in]{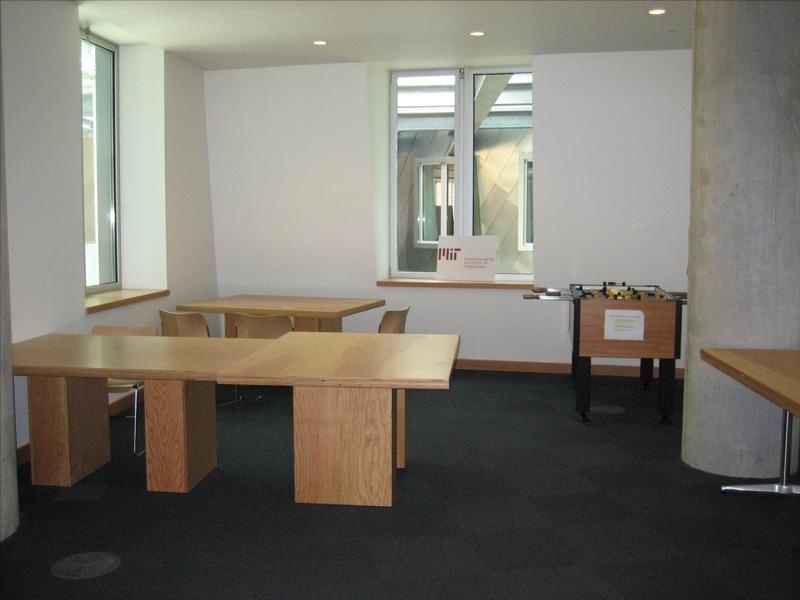} & \includegraphics[width=1.2in, height=0.8in]{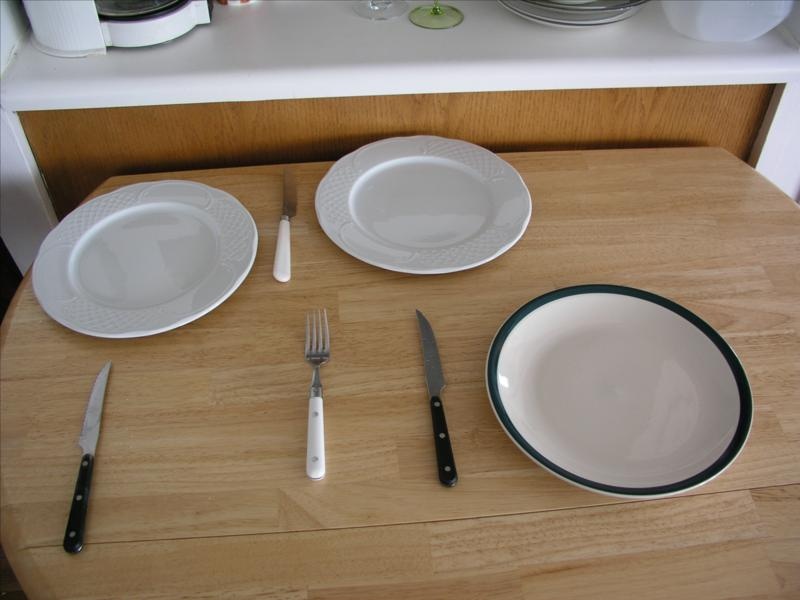} & 
         \includegraphics[width=1.2in, height=0.8in]{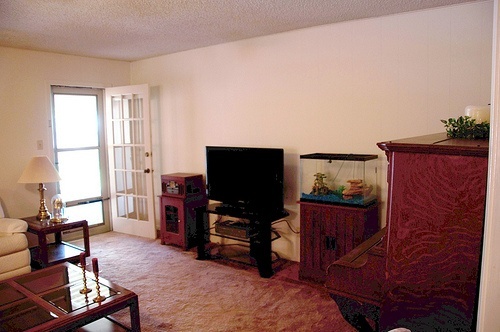} \\
         ours & \includegraphics[width=1.2in, height=0.8in]{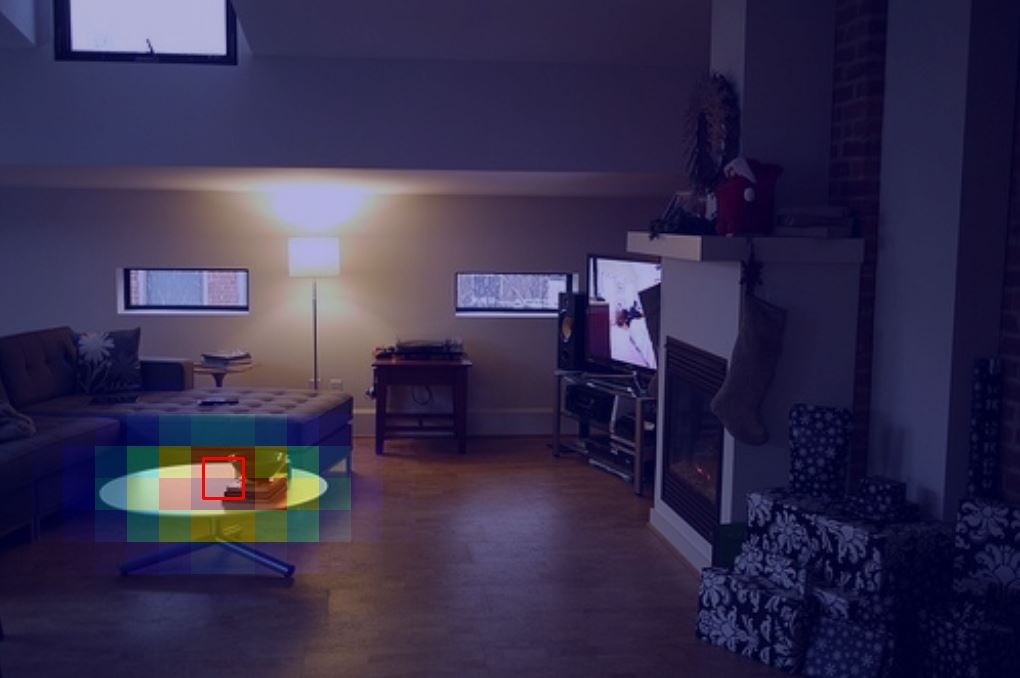} & \includegraphics[width=1.2in, height=0.8in]{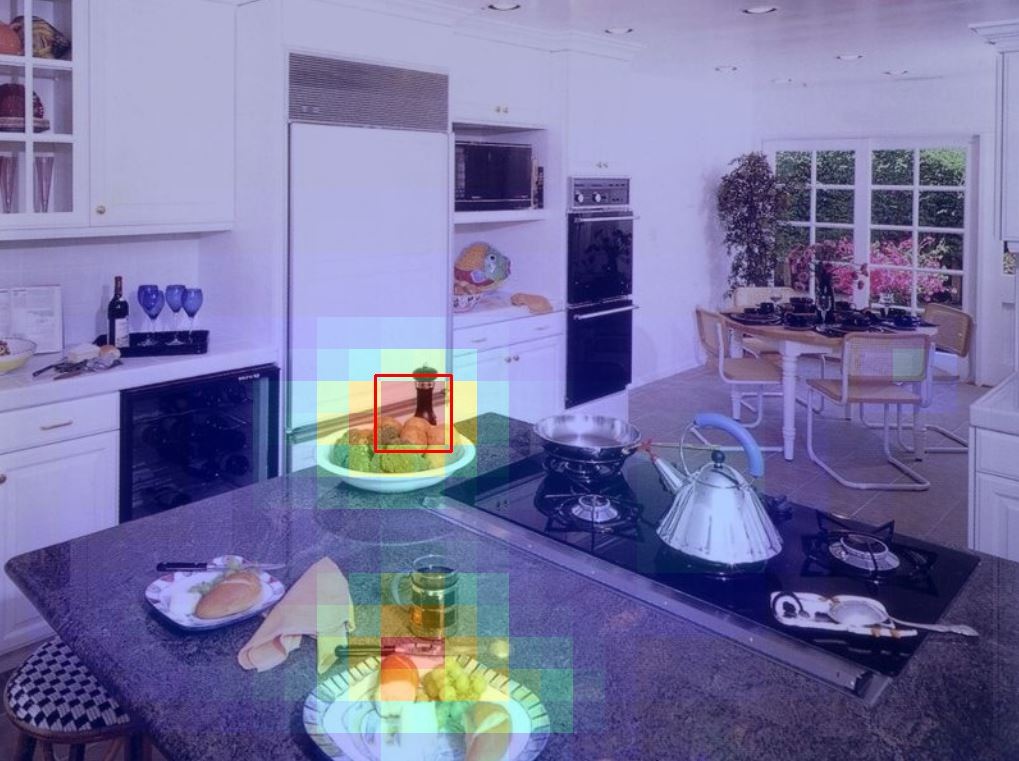} & \includegraphics[width=1.2in, height=0.8in]{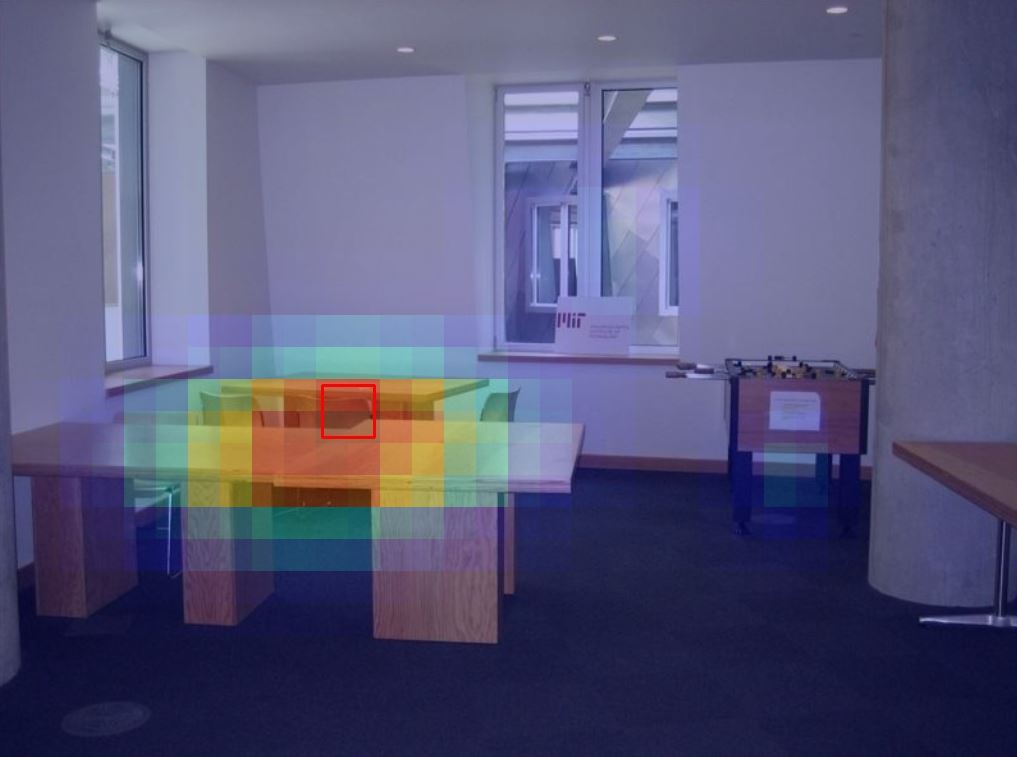} & \includegraphics[width=1.2in, height=0.8in]{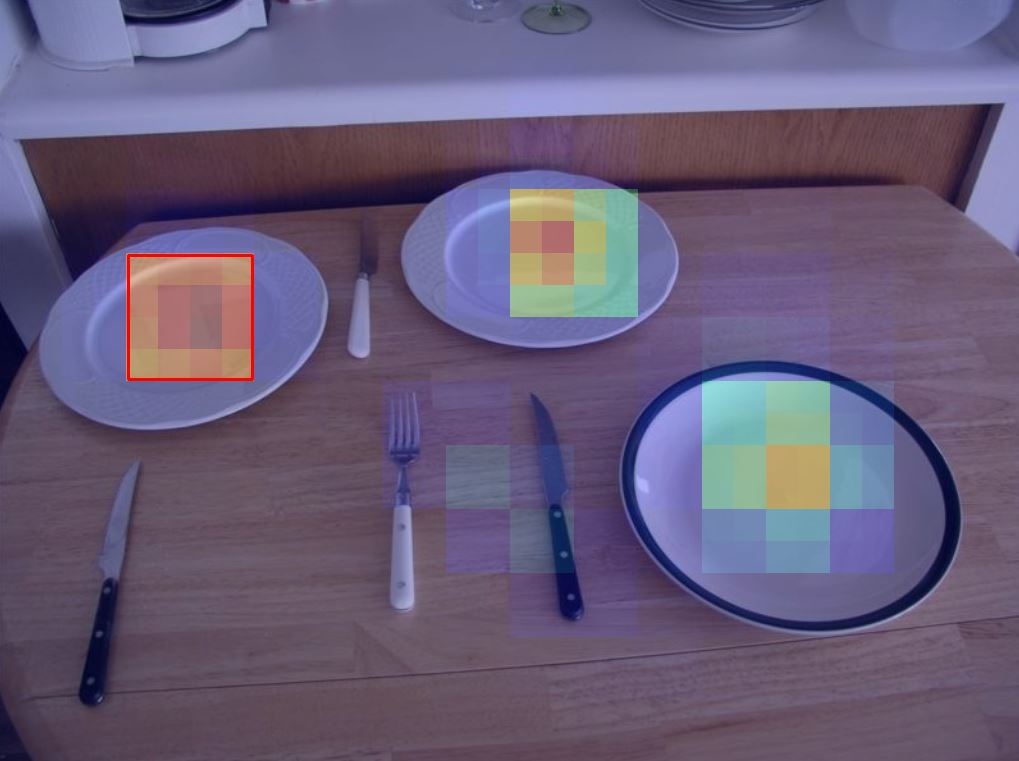} & \includegraphics[width=1.2in, height=0.8in]{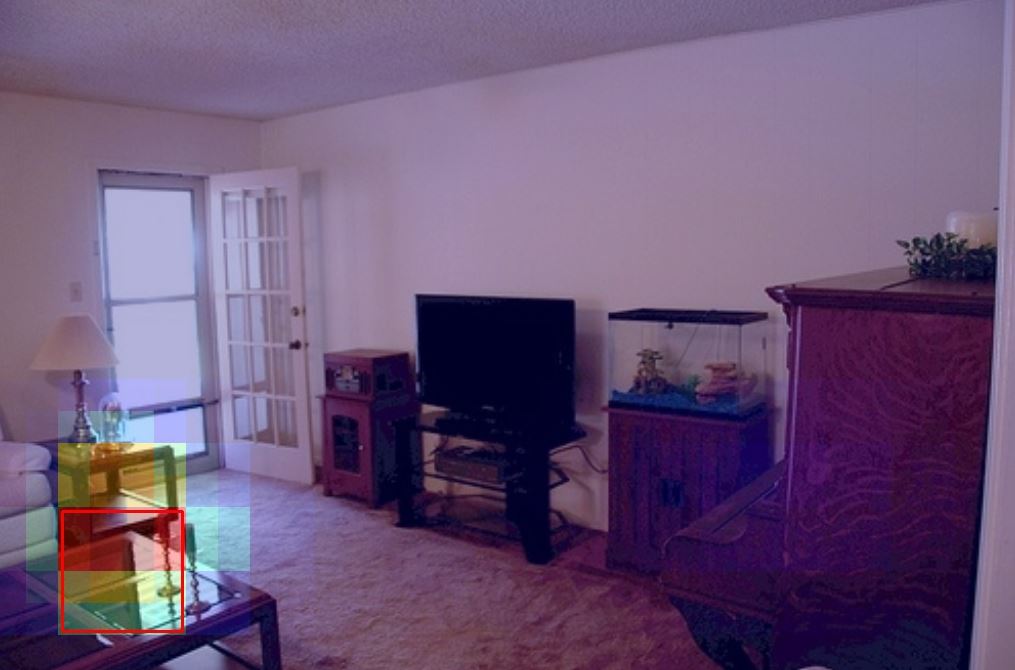} \\
         \cite{DBLP:journals/corr/TanBCOB17} & \includegraphics[width=1.2in, height=0.8in]{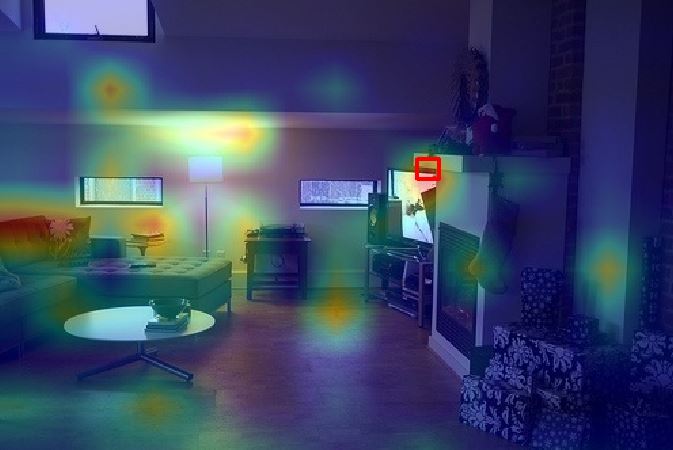} & \includegraphics[width=1.2in, height=0.8in]{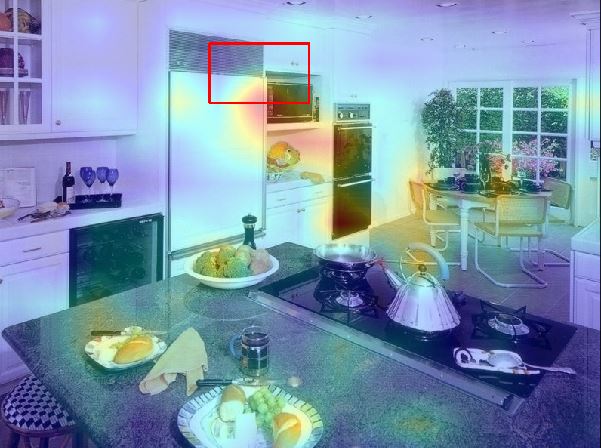} & \includegraphics[width=1.2in, height=0.8in]{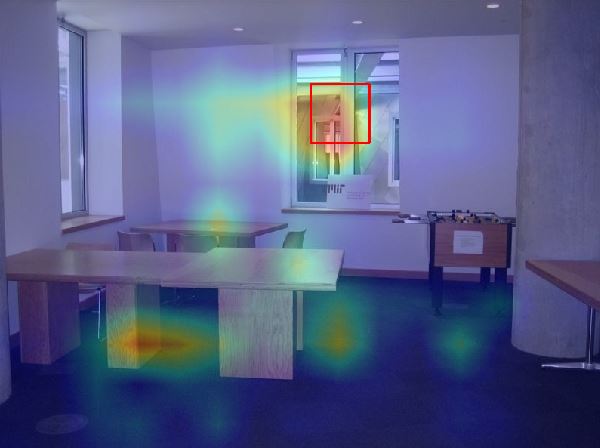} & \includegraphics[width=1.2in, height=0.8in]{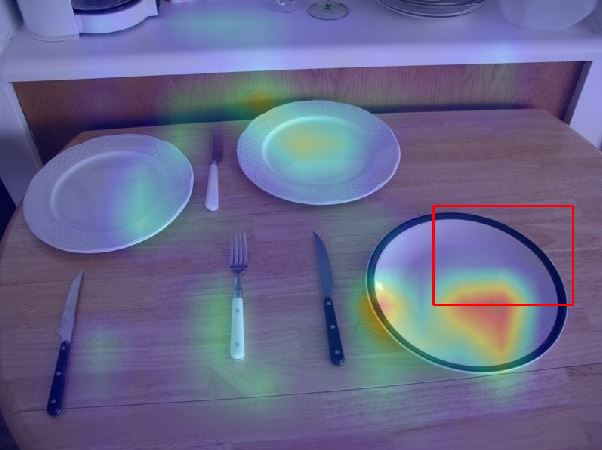} & \includegraphics[width=1.2in, height=0.8in]{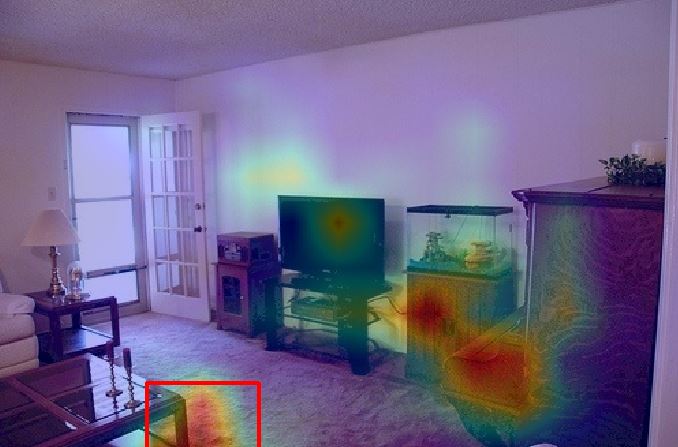} \\
         & \textit{cup} & \textit{spoon} & \textit{apple} & \textit{cake} & \textit{laptop} \\\\
         
         input & \includegraphics[width=1.2in, height=0.8in]{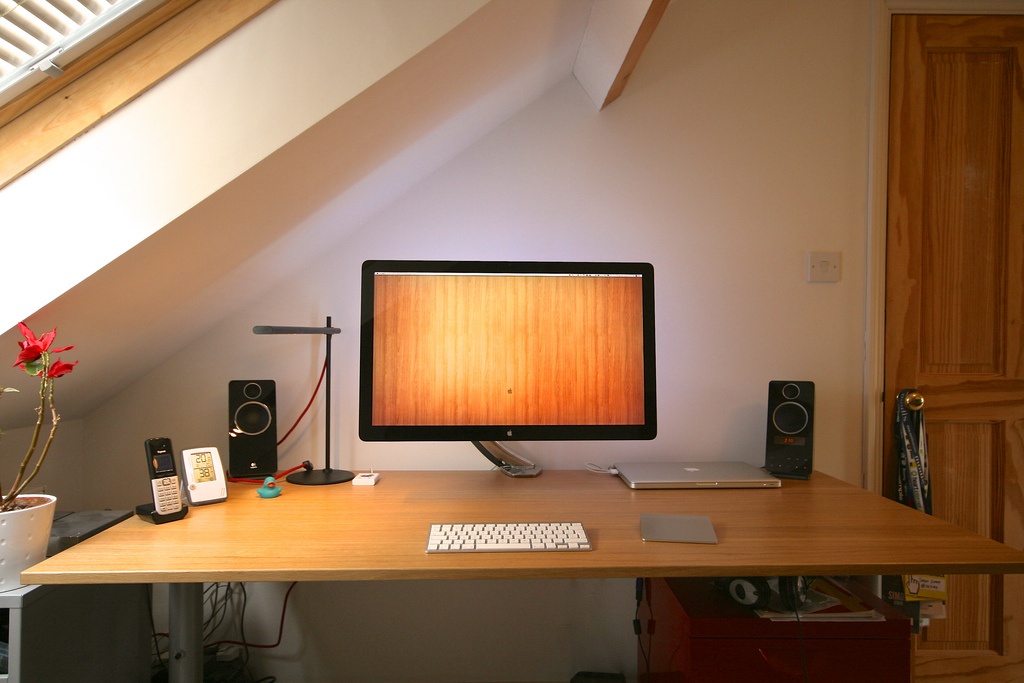} & \includegraphics[width=1.2in, height=0.8in]{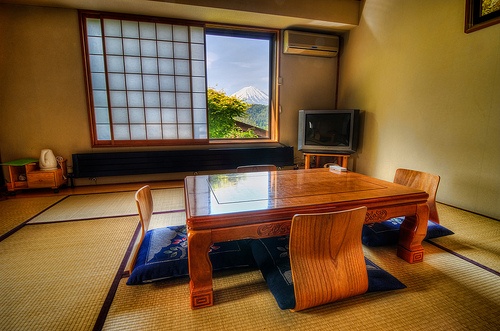} & 
         \includegraphics[width=1.2in, height=0.8in]{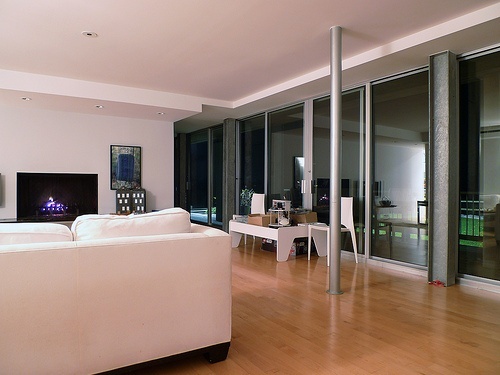} & \includegraphics[width=1.2in, height=0.8in]{assets/book1.jpg} & 
         \includegraphics[width=1.2in, height=0.8in]{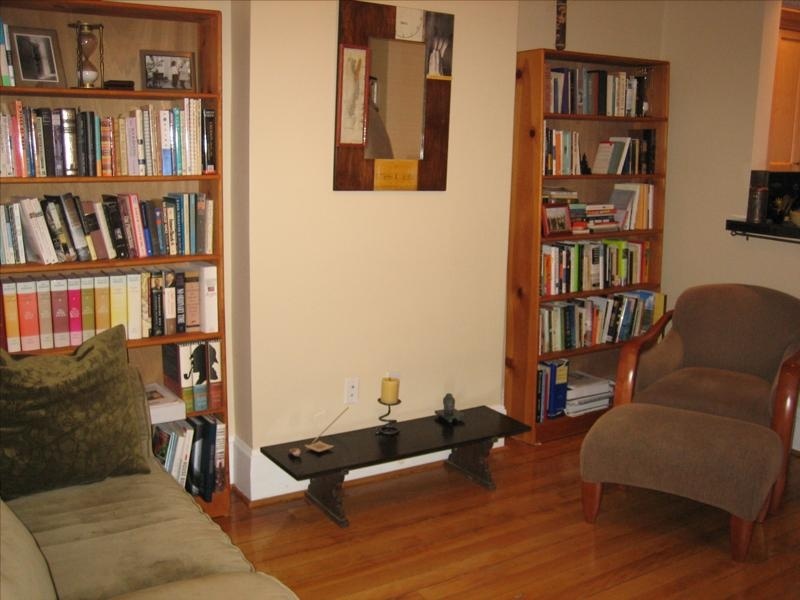} \\
         ours & \includegraphics[width=1.2in, height=0.8in]{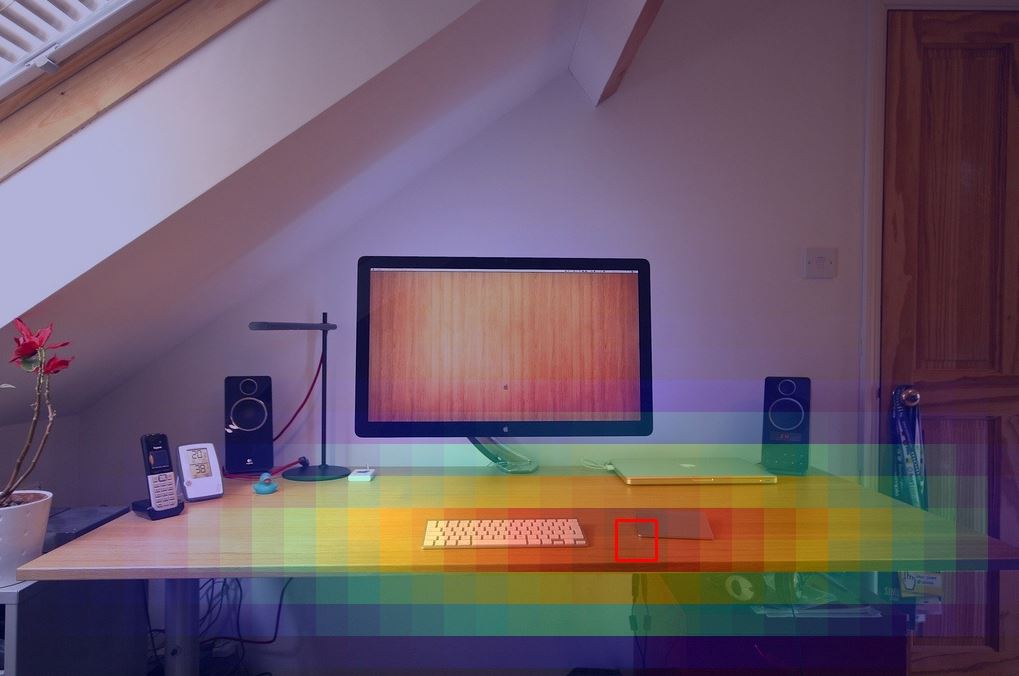} & \includegraphics[width=1.2in, height=0.8in]{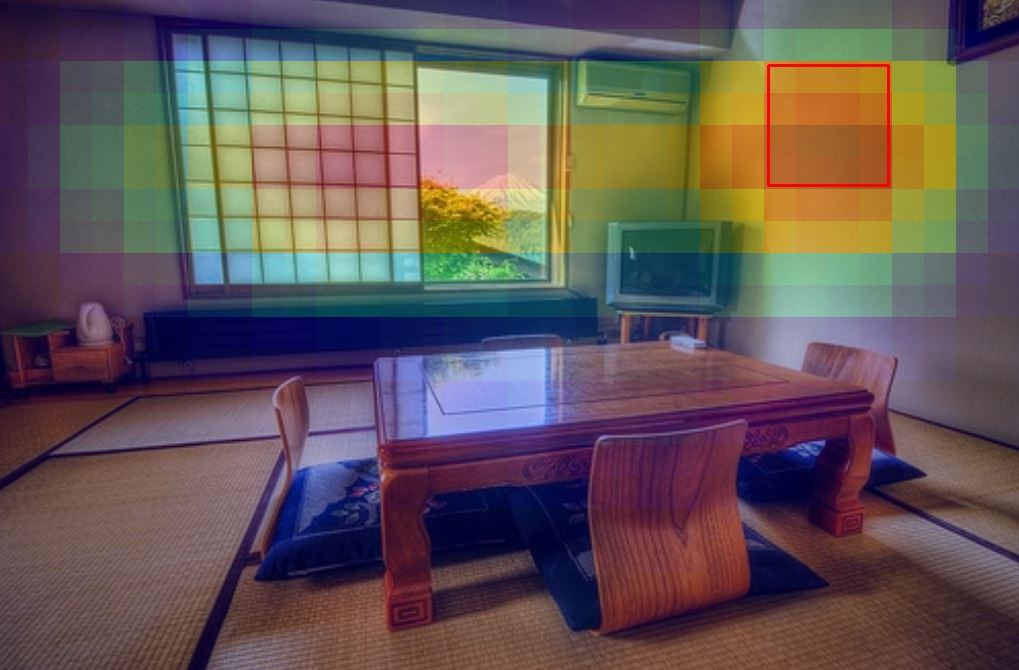} & \includegraphics[width=1.2in, height=0.8in]{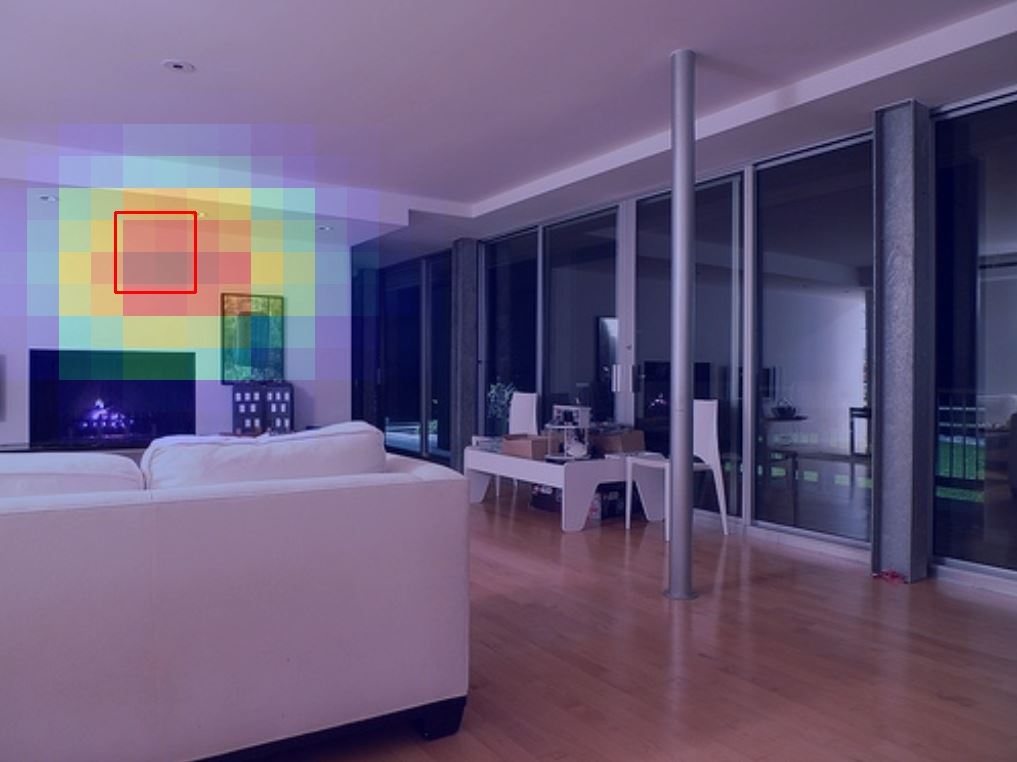} & \includegraphics[width=1.2in, height=0.8in]{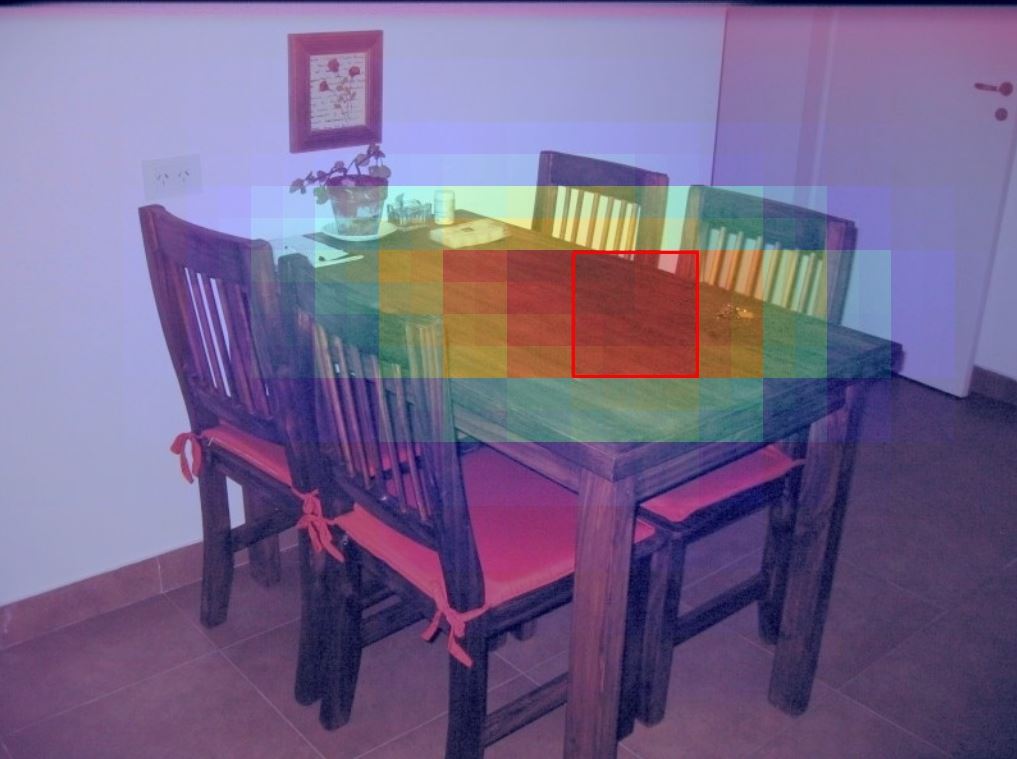} & \includegraphics[width=1.2in, height=0.8in]{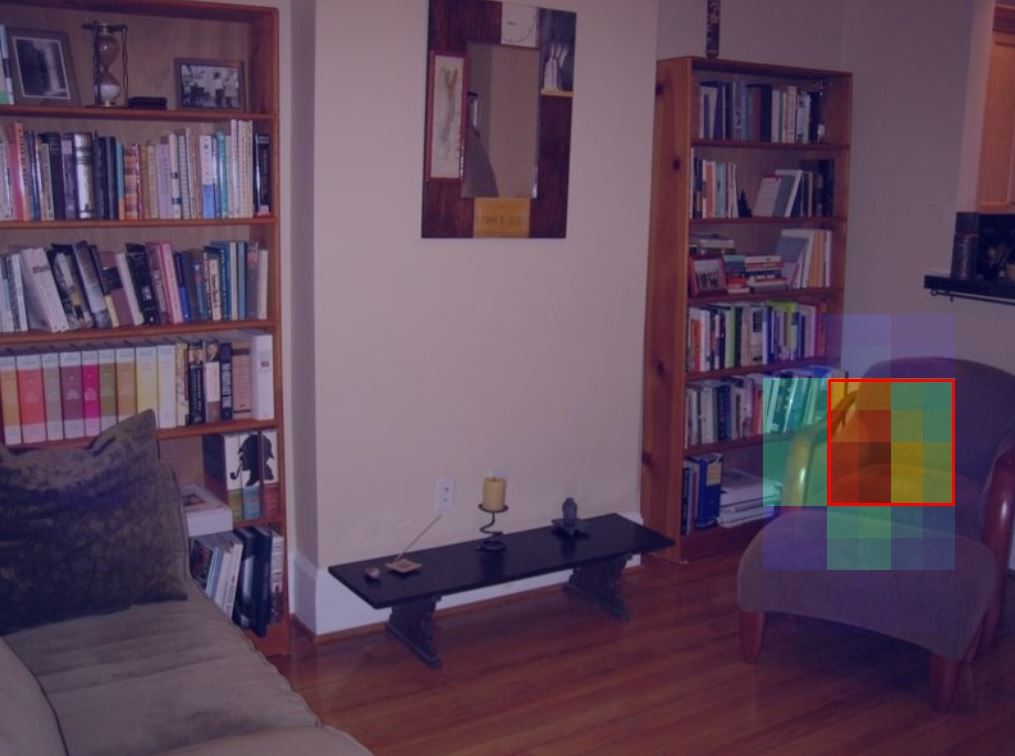} \\
         \cite{DBLP:journals/corr/TanBCOB17} & \includegraphics[width=1.2in, height=0.8in]{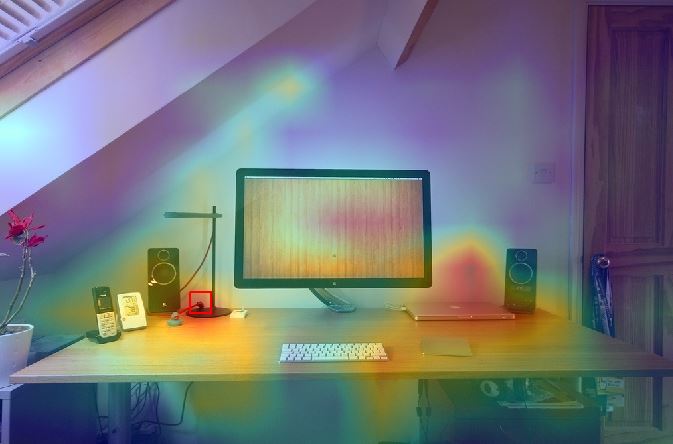} & \includegraphics[width=1.2in, height=0.8in]{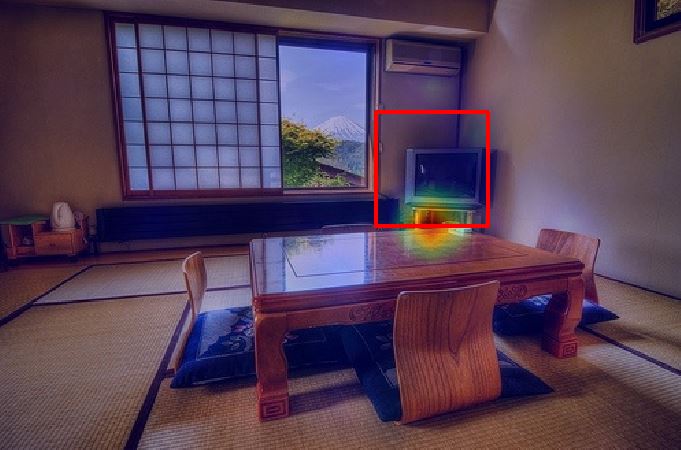} & \includegraphics[width=1.2in, height=0.8in]{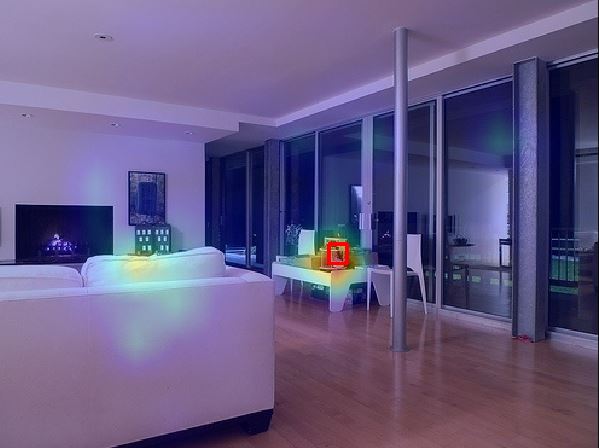} & \includegraphics[width=1.2in, height=0.8in]{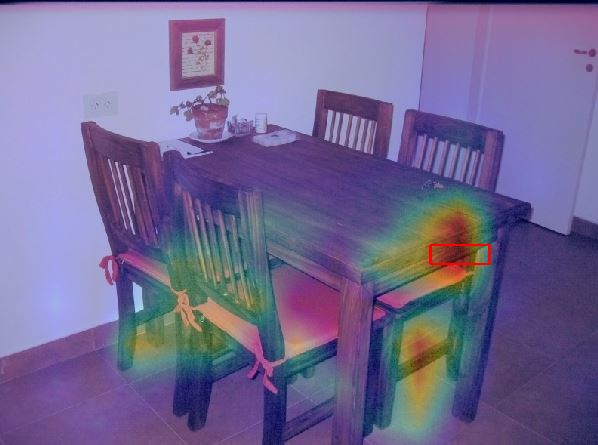} & \includegraphics[width=1.2in, height=0.8in]{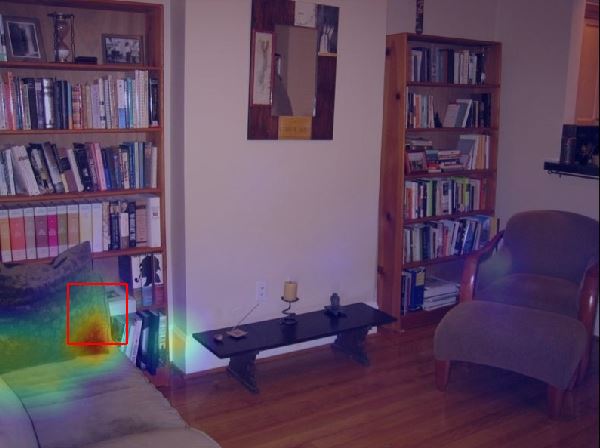} \\
         & \textit{mouse} & \textit{tv} & \textit{clock} & \textit{book} & \textit{pillow} \\
    \end{tabular}
    \caption{Qualitative comparison against \cite{DBLP:journals/corr/TanBCOB17} on bounding box prediction. The first row is the original images, the second row is our results, the third row is the results of \cite{DBLP:journals/corr/TanBCOB17}. The inserted object for each image is labeled at the bottom of each column.}
    \label{table:bbox_pred}
\end{table*}

\section{Conclusion}

We propose a novel research topic, \textit{dual recommendation for object insertion}, and build an unsupervised algorithm that exploits object-level context. We establish a new test dataset and design task-specific metrics for automatic quantitative evaluation. We outperform existing baselines on all subtasks under a unified framework, as evidenced by both quantitative and qualitative results. Future work includes incorporation of high-dimensional image features, or larger datasets that is able to fully drive the training of neural networks.

\section*{Acknowledgment}
This work was supported by the National Key R\&D Program (No. 2017YFB1002604), the National Natural Science Foundation of China (No. 61772298 and No. 61521002), a Research Grant of Beijing Higher
Institution Engineering Research Center, and the Tsinghua-Tencent Joint Laboratory for Internet Innovation Technology.

{\small
\bibliographystyle{IEEEtran}
\bibliography{references}
}

\begin{IEEEbiography}[{\includegraphics[width=1in,height=1.25in,clip,keepaspectratio]{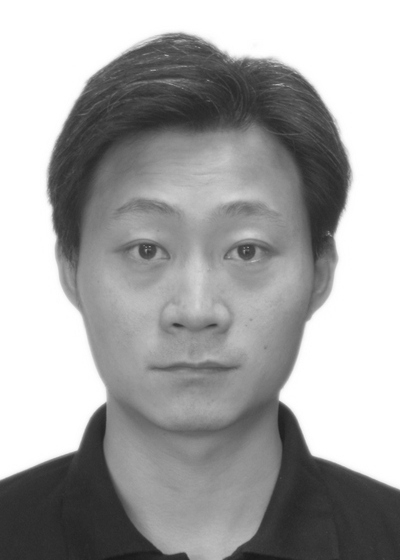}}]{Song-Hai Zhang}
received the Ph.D. degree from Tsinghua University, Beijing, China, in 2007. He is currently an associate professor of computer science at Tsinghua University. His research interests include image and video processing as well as geometric computing. He is the corresponding author.
\end{IEEEbiography}

\begin{IEEEbiography}[{\includegraphics[width=1in,height=1.25in,clip,keepaspectratio]{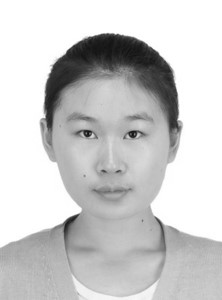}}]{Zhengping Zhou}
is an undergraduate student in the Department of Computer Science and Technology, Tsinghua University. She is expected to receive her bachelor's degree in computer science from the same university in 2019. Her research interests include image processing and computer graphics.
\end{IEEEbiography}

\begin{IEEEbiography}[{\includegraphics[width=1in,height=1.25in,clip,keepaspectratio]{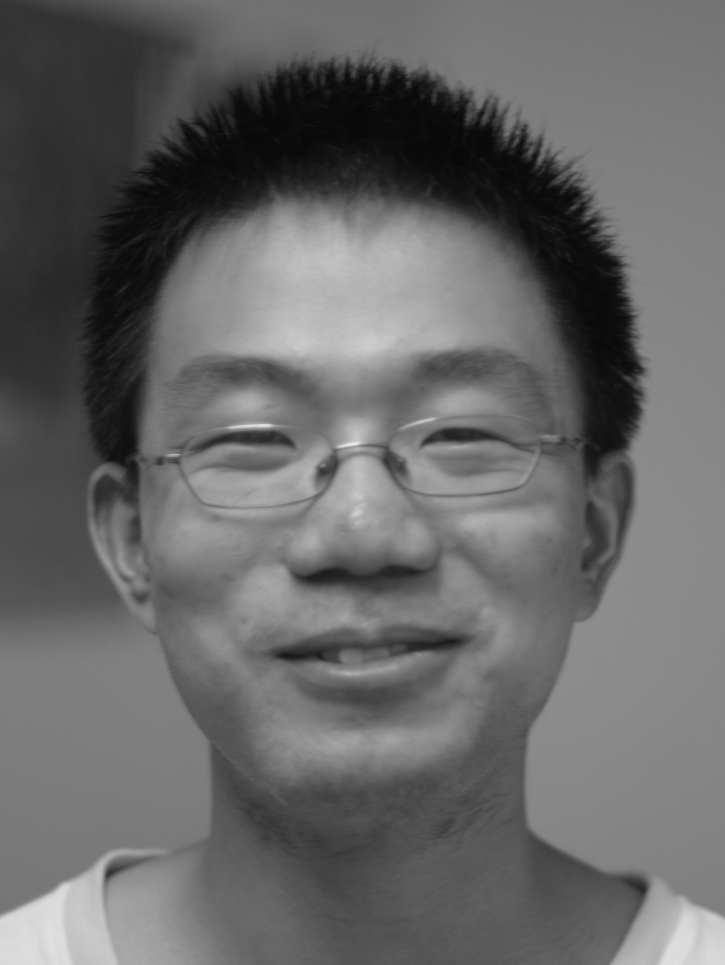}}]{Bin Liu}
is a PhD student in the Department of Computer Science and Technology, Tsinghua University. He received his bachelor's degree in computer science from the same university in 2013. His research interests include image and video editing.
\end{IEEEbiography}

\begin{IEEEbiography}[{\includegraphics[width=1in,height=1.25in,clip,keepaspectratio]{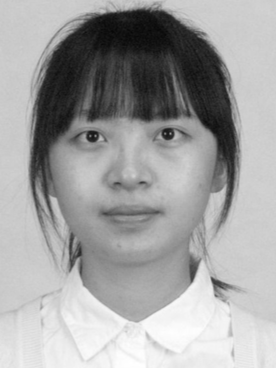}}]{Xin Dong}
is a Master student in the Department of Computer Science and Technology, Tsinghua University. She received her bachelor's degree in computer science from the same university in 2016. Her research interests include image understanding.
\end{IEEEbiography}

\begin{IEEEbiography}[{\includegraphics[width=1in,height=1.25in,clip,keepaspectratio]{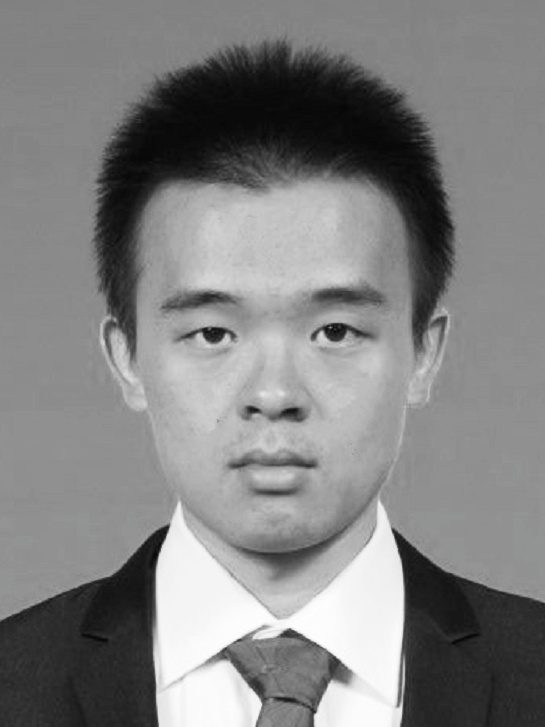}}]{Dun Liang}
is a PhD student in the Department of Computer Science and Technology, Tsinghua University. He received his bachelor's degree in computer science from the same university in 2016. His research interests include image processing and neural networks.
\end{IEEEbiography}

\begin{IEEEbiography}[{\includegraphics[width=1in,height=1.25in,clip,keepaspectratio]{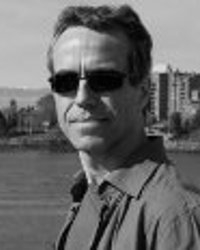}}]{Peter Hall}
is an associate professor in the Department of Computer Science at the University of Bath. He is also the director of the Media Technology Research Centre, Bath. He founded to vision, video, and graphics network of excellence in the United Kingdom, and has served on the executive committee of the British Machine Vision Conference since 2003. He has published extensively in Computer Vision, especially where it interfaces with Computer Graphics. More recently he is developing an interest in robotics. 
\end{IEEEbiography}

\begin{IEEEbiography}[{\includegraphics[width=1in,height=1.25in,clip,keepaspectratio]{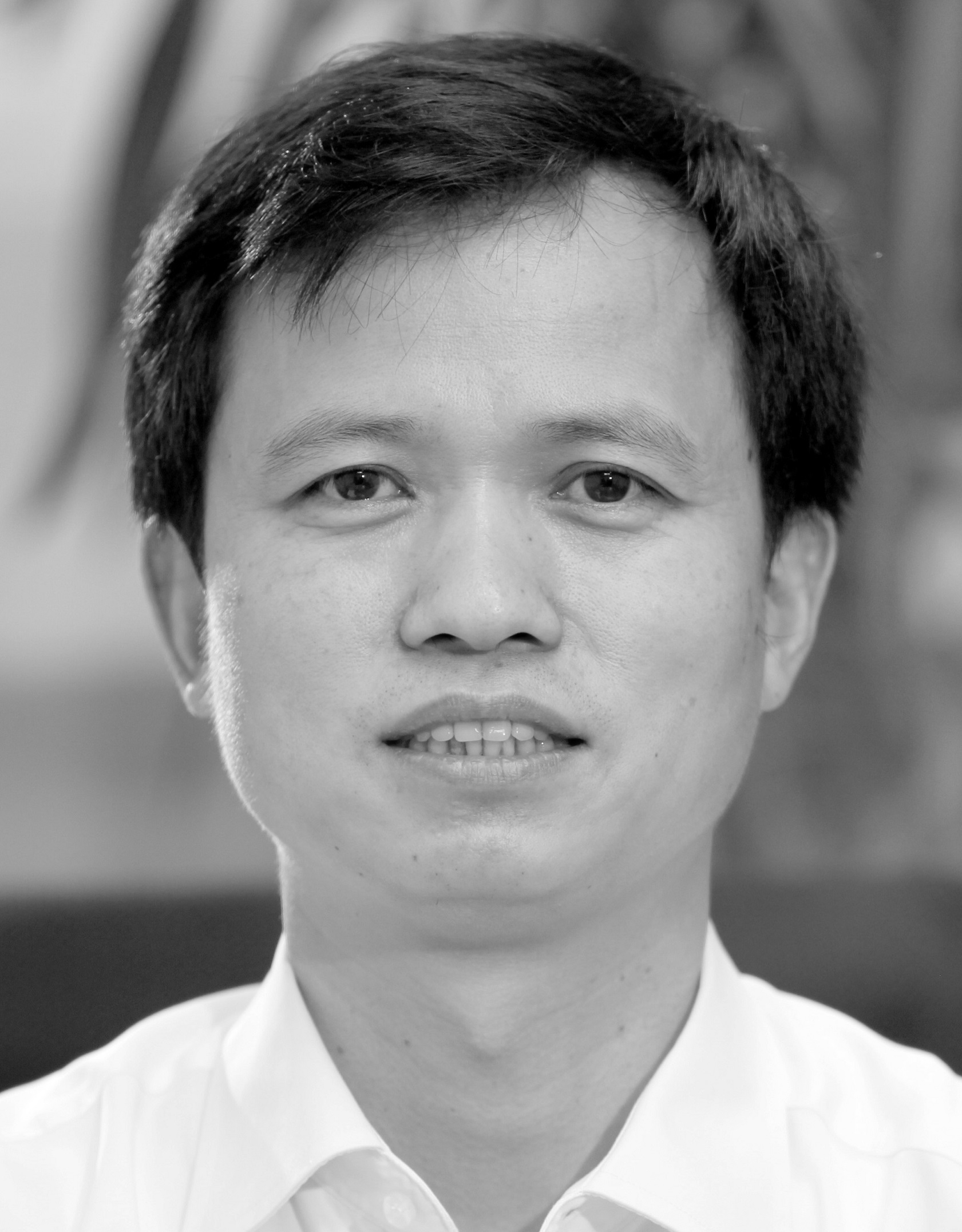}}]{Shi-Min Hu}
received the Ph.D. degree from Zhejiang University, in 1996. He is currently a Professor with the Department of Computer Science and Technology, Tsinghua University, Beijing. His research interests include digital geometry processing, video processing, computer animation and computer-aided geometric design. He is the Editor-in-Chief of Computational Visual Media, and on the Editorial Board of several journals, including the IEEE Transactions on Visualization and Computer Graphics, and Computer Aided Design and Computer and Graphics.
\end{IEEEbiography}

\end{document}